\documentclass[twoside]{article}

\usepackage[accepted]{aistats2024}

\usepackage[utf8]{inputenc}
\usepackage[T1]{fontenc}
\usepackage{hyperref}
\usepackage{url}
\usepackage{booktabs}
\usepackage{nicefrac}
\usepackage{microtype}
\usepackage{xcolor}
\usepackage{amsfonts}
\usepackage{amsmath}
\usepackage{amssymb}
\usepackage{mathtools}
\usepackage{amsthm}
\usepackage{graphicx}
\usepackage{wrapfig}
\usepackage{bm}
\usepackage{cleveref}
\usepackage{float}
\usepackage{tabularx,ragged2e,booktabs,caption}
\usepackage{graphicx,subcaption}
\usepackage{natbib}
\usepackage{listings}

\definecolor{mydarkblue}{rgb}{0,0.08,0.45}
\hypersetup{ 
    colorlinks=true,
    linkcolor=mydarkblue,
    citecolor=mydarkblue,
    filecolor=mydarkblue,
    urlcolor=mydarkblue,
}

\usepackage[textsize=tiny]{todonotes}

\definecolor{codegreen}{rgb}{0,0.6,0}
\definecolor{codegray}{rgb}{0.5,0.5,0.5}
\definecolor{codepurple}{rgb}{0.58,0,0.82}
\definecolor{backcolour}{rgb}{0.95,0.95,0.92}

\lstdefinestyle{mystyle}{
    backgroundcolor=\color{backcolour},   
    commentstyle=\color{codegreen},
    keywordstyle=\color{magenta},
    numberstyle=\tiny\color{codegray},
    stringstyle=\color{codepurple},
    basicstyle=\ttfamily\footnotesize,
    breakatwhitespace=false,         
    breaklines=true,                 
    captionpos=b,                    
    keepspaces=true,                 
    numbers=left,                    
    numbersep=5pt,                  
    showspaces=false,                
    showstringspaces=false,
    showtabs=false,                  
    tabsize=2
}

\begin{document}

%
\runningtitle{Differentiable Rendering with Reparameterized Volume Sampling}

%
\runningauthor{Morozov, Rakitin, Desheulin, Vetrov, Struminsky}

\twocolumn[

\aistatstitle{Differentiable Rendering with
              Reparameterized Volume Sampling}

\aistatsauthor{ Nikita Morozov\And Denis Rakitin \And  Oleg Desheulin}

\aistatsaddress{HSE Unversity \And  HSE University \And  HSE University}

\aistatsauthor{Dmitry Vetrov$^\ast$ \And Kirill Struminsky }

\aistatsaddress{Constructor University, Bremen \And HSE Unversity}]

\begin{abstract}
In view synthesis, a neural radiance field approximates underlying density and radiance fields based on a sparse set of scene pictures.
To generate a pixel of a novel view, it marches a ray through the pixel and computes a weighted sum of radiance emitted from a dense set of ray points.
This rendering algorithm is fully differentiable and facilitates gradient-based optimization of the fields.
However, in practice, only a tiny opaque portion of the ray contributes most of the radiance to the sum.
We propose a simple end-to-end differentiable sampling algorithm based on inverse transform sampling. It generates samples according to the probability distribution induced by the density field and picks non-transparent points on the ray. We utilize the algorithm in two ways. First, we propose a novel rendering approach based on Monte Carlo estimates. This approach allows for evaluating and optimizing a neural radiance field with just a few radiance field calls per ray. Second, we use the sampling algorithm to modify the hierarchical scheme proposed in the original NeRF work. We show that our modification improves reconstruction quality of hierarchical models, at the same time simplifying the training procedure by removing the need for auxiliary proposal network losses.
\end{abstract}

\begin{figure}[H]
\centering
\includegraphics[width=0.9\linewidth]{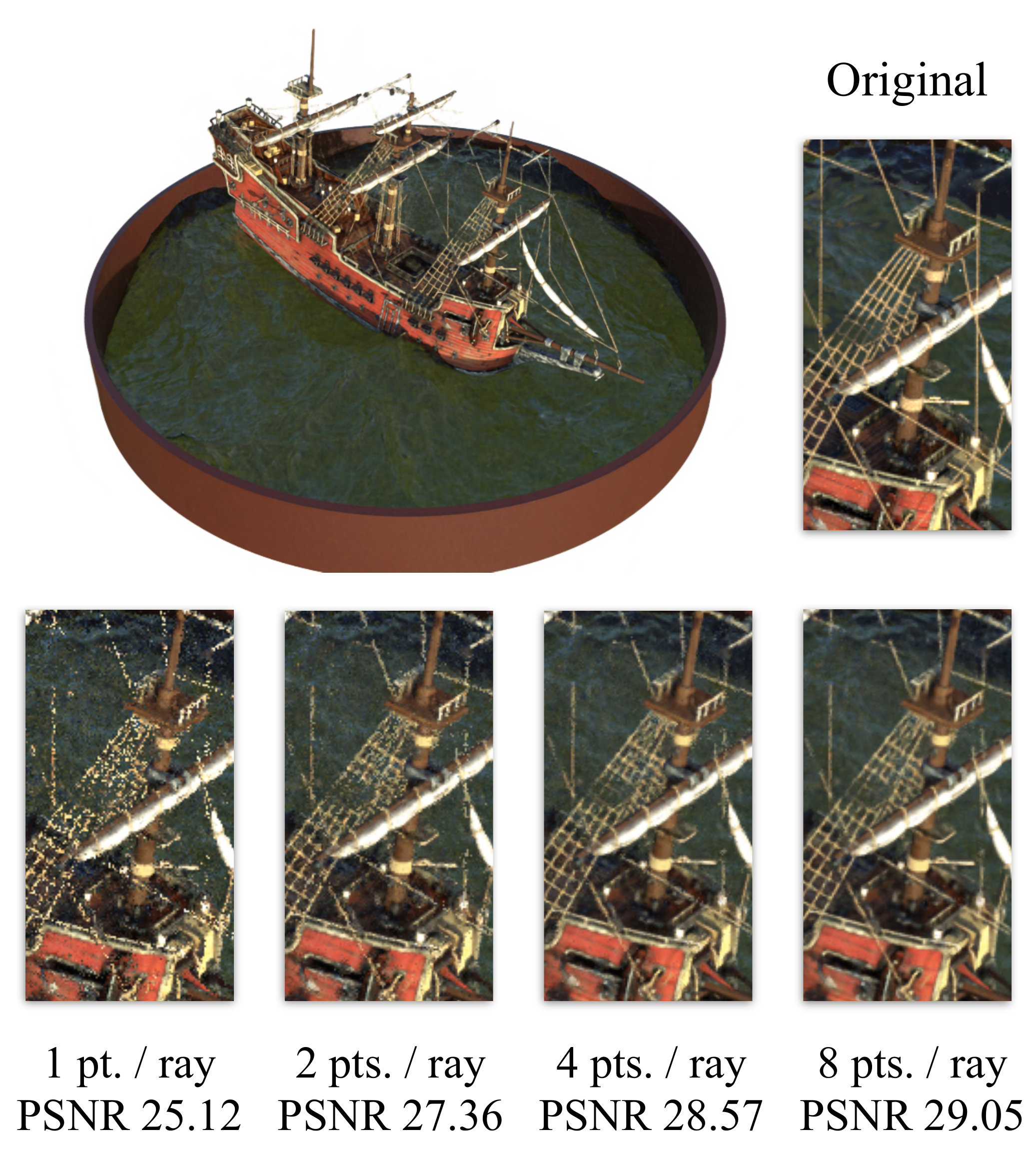}
\caption{Novel views of a ship generated with the proposed Monte Carlo radiance estimates. For each ray we estimate density and then compute radiance at a few ray points generated using the ray density. As the above images indicate, render quality gradually improves with the number of ray samples, without visible artifacts at eight points per ray.}
\label{fig:summary}
\end{figure}

\section{INTRODUCTION}
\label{gen_inst}

Given a set of scene pictures with corresponding camera positions, novel view synthesis aims to generate pictures of the same scene from new camera positions.
Recently, learning-based approaches have led to significant progress in this area.
As an early instance, neural radiance fields (NeRF) by~\cite{mildenhall2020nerf} represent a scene via a density field and a radiance (color) field parameterized with a multilayer perceptron (MLP).
Using a differentiable volume rendering algorithm~\citep{max1995optical} with MLP-based fields to produce images, they minimize the discrepancy between the output images and a set of reference images to learn a scene representation.

In particular, NeRF generates an image pixel by casting a ray from a camera through the pixel and aggregating the radiance at each ray point with weights induced by the density field.
Each term involves a costly neural network query, and the model has a trade-off between rendering quality and computational load.
In this work, we revisit the formula for the aggregated radiance computation and propose a novel approximation based on Monte Carlo methods. We compute our approximation in two stages.
In the first stage, we march through the ray to estimate density.
In the second stage, we construct a Monte Carlo color approximation using the density to pick points along the ray.
The resulting estimate is fully differentiable and can act as a drop-in replacement for the standard rendering algorithm used in NeRF.
Fig.~\ref{fig:summary} illustrates the estimates for a varying number of samples.
Compared to the standard rendering algorithm, the second stage of our algorithm avoids redundant radiance queries and can potentially reduce computation during training and inference.

Furthermore, we show that the sampling algorithm used in our Monte Carlo estimate is applicable to the hierarchical sampling scheme in NeRF. Similar to our work, the hierarchical scheme uses inverse transform sampling to pick points along a ray.
The corresponding distribution is tuned using an auxiliary training task.
In contrast, we derive our algorithm from a different perspective and obtain the inverse transform sampling for a slightly different distribution.
With our algorithm, we were able to train NeRF end-to-end without the auxiliary task and improve the reconstruction quality.
We achieve this by back-propagating the gradients through the sampler, and show that the original sampling algorithm fails to achieve similar quality in the same setup.

Below, Section~\ref{sec:nerf-recap} gives a recap of neural radiance fields. Then we proceed to the main contributions of our work in Section~\ref{sec:theory}, namely the rendering algorithm fueled by Monte Carlo estimates and the novel sampling procedure. In Section~\ref{sec:related-work} we discuss related work. In Subsection~\ref{sec:nerf}, we use our sampling algorithm to improve the hierarchical sampling scheme proposed for training NeRF.
Finally, in Subsection~\ref{sec:dvgo} we apply the proposed Monte Carlo estimate to replace the standard rendering algorithm.
With an efficient neural radiance field architecture, our algorithm decreases time per training iteration at the cost of reduced reconstruction quality. We also show that our Monte Carlo estimate can be used during inference of a pre-trained model with no additional fine-tuning needed, and it can achieve better reconstruction quality at the same speed in comparison to the standard algorithm. Our source code is available at \url{https://github.com/GreatDrake/reparameterized-volume-sampling}.

\section{NEURAL RADIANCE FIELDS}\label{sec:nerf-recap}
Neural radiance fields represent 3D scenes with a non-negative scalar density field $\sigma: \mathbb R^3 \rightarrow \mathbb R^+$ and a vector radiance field $c: \mathbb R^3 \times \mathbb R^3 \rightarrow \mathbb R^3$. Scalar field $\sigma$ represents volume density at each spatial location $\bm{x}$, and $c(\bm{x}, \bm{d})$ returns the light emitted from spatial location $\bm{x}$ in direction $\bm{d}$ represented as a normalized three dimensional vector. 

For novel view synthesis, NeRF~\citep{mildenhall2020nerf} adapts a volume rendering algorithm that computes pixel color $C(\bm{r})$ as the expected radiance for a ray $\bm{r} = \bm{o} + t \bm{d}$ passing through a pixel from origin $\bf{o} \in \mathbb R^3$ in a direction $\bf{d} \in \mathbb R^3$. For ease of notation, we will denote density and radiance restricted to a ray $\bm{r}$ as
\begin{align}
\sigma_{\bm{r}}(t) := \sigma(\bm{o} + t \bm{d}) \text{\;and\;}
c_{\bm{r}}(t) := c(\bm{o} + t \bm{d}, \bm{d}).
\end{align}
With that in mind, the expected radiance along ray $\bm{r}$ is given as 
\begin{equation}\label{eq:expected_color}
C(\bm{r}) = \int_{t_n}^{t_f} p_{\bm{r}}(t) c_{\bm{r}}(t) \mathrm{d} t,
\end{equation}
where
\begin{equation}\label{eq:density_field_dist}
p_{\bm{r}}(t) := \sigma_{\bm{r}}(t) \exp{\left(- \int_{t_n}^{t} \sigma_{\bm{r}} (s) \mathrm{d} s \right)}.
\end{equation}
Here, $t_n$ and $t_f$ are {\it near} and {\it far} ray boundaries, and $p_{\bm{r}}(t)$ is an unnormalized probability density function of a random variable $\textnormal{t}$ on a ray $\bm{r}$. Intuitively, $\textnormal{t}$ is the location on a ray where the portion of light coming into the point $\bm{o}$ was emitted.

To approximate the nested integrals in Eq.~\ref{eq:expected_color}, \cite{max1995optical} proposed to replace fields $\sigma_{\bm{r}}$ and $c_{\bm{r}}$ with a piecewise approximation on a grid $t_n = t_0 < t_1 < \dots < t_m = t_f$ and compute the formula in Eq.~\ref{eq:expected_color} analytically for the approximation. In particular, a piecewise constant approximation with density $\sigma_i$ and radiance $c_i$ within $i$-th bin $[t_{i + 1}, t_{i}]$ of width $\delta_i = t_{i + 1} - t_i$ yields formula
\begin{equation}\label{eq:grid_color_approximation}
\hat{C}(\bm{r}) = \sum_{i=1}^m w_i c_i, 
\end{equation}
where the weights are given by
\begin{equation}\label{eq:nerf_weights}
w_i = (1 - \exp(-\sigma_i \delta_i)) \exp \left( - \sum_{j=1}^{i-1} \sigma_j \delta_j \right).
\end{equation}
Importantly, Eq.~\ref{eq:grid_color_approximation} is fully differentiable and can be used as a part of a gradient-based learning pipeline. To reconstruct a scene NeRF runs a gradient based optimizer to minimize MSE between the predicted color and the ground truth color averaged across multiple rays and multiple viewpoints.

While the above approximation works in practice, it involves multiple evaluations of $c$ and $\sigma$ along a dense grid. Besides that, a ray typically intersects a solid surface at some point $t \in [t_n, t_f]$. In this case, probability density $p_{\bm{r}}(t)$ will concentrate its mass near $t$ and, as a result, most of the terms in Eq.~\ref{eq:grid_color_approximation} will make a negligible contribution to the sum. To approach this problem, NeRF employs a hierarchical sampling scheme. Two networks are trained simultaneously: coarse (or proposal) and fine. Firstly, the coarse network is evaluated on a uniform grid of $N_c$ points and a set of weights $w_i$ is calculated as in Eq.~\ref{eq:nerf_weights}. Normalizing these weights produces a piecewise constant PDF along the ray. Then $N_f$ samples are drawn from this distribution and the union of the first and second sets of points is used to evaluate the fine network and compute the final color estimation. The coarse network is also trained to predict ground truth colors, but the color estimate for the coarse network is calculated only using the first set of $N_c$ points.

\section{REPARAMETERIZED VOLUME SAMPLING AND RADIANCE ESTIMATES}
\label{sec:theory}

\subsection{Reparameterized Expected Radiance Estimates}
\label{sec:mc_est}

Monte Carlo methods give a natural way to approximate the expected color. For example, given $k$ i.i.d. samples $t_1, \dots, t_k \sim p_{\bm{r}}(t)$ and the normalizing constant $y_f := \int_{t_n}^{t_f} p_{\bm{r}} (t) \mathrm d t$, the sum
\begin{equation}\label{eq:mc_color_approximation}
\hat{C}_{MC}(\bm{r}) = \frac{y_f}{k} \sum_{i=1}^k c_{\bm{r}}(t_i)
\end{equation} 
is an unbiased estimate of the expected radiance in Eq.~\ref{eq:expected_color}.
Moreover, samples $t_1, \dots, t_k$ belong to high-density regions of $p_{\bm{r}}$ by design, thus for a degenerate density $p_{\bm{r}}$ even a few samples would provide an estimate with low variance.
Importantly, unlike the approximation in Eq.~\ref{eq:grid_color_approximation}, the Monte Carlo estimate depends on scene density $\sigma$ implicitly through sampling algorithm and requires a custom gradient estimate for the parameters of $\sigma$.
We propose a principled end-to-end differentiable algorithm to generate samples from $p_{\bm{r}}(t)$.

Our solution is primarily inspired by the reparameterization trick~\citep{kingma2014adam, rezende2014stochastic}. We change the variable in Eq.~\ref{eq:expected_color}.
For ${F_{\bm{r}}(t) := 1 - \exp{\left(-\int_{t_n}^{t} \sigma_{\bm{r}}(s) \mathrm d s \right)}}$ and $y := F_{\bm{r}}(t)$ we rewrite
\begin{align}
C(\bm{r}) &= \int_{t_n}^{t_f} c_{\bm{r}}(t) p_{\bm{r}}(t) \mathrm d t \label{eq:y_reparameterization_1}\\
          &= \int_{y_n}^{y_f} c_{\bm{r}}(F_{\bm{r}}^{-1}(y)) \mathrm d y \label{eq:y_reparameterization_2}\\
          &= \int_{0}^{1} y_f c_{\bm{r}}(F_{\bm{r}}^{-1}( y_f u)) \mathrm d u. \label{eq:y_reparameterization_3}
\end{align}
The integral boundaries are $y_n := F_{\bm{r}}(t_n) = 0$ and $y_f := F_{\bm{r}}(t_f)$.
Function $F_{\bm{r}}(t)$ acts as the cumulative distribution function of the variable $\textnormal{t}$ with a single exception that, in general, $F_{\bm{r}}(t_f) \neq 1$.
In volume rendering, $F_{\bm{r}}(t)$ is called opacity function with $y_f$ being equal to overall pixel opaqueness.
After the first change of variables in Eq.~\ref{eq:y_reparameterization_2}, the integral boundaries depend on opacity $F_{\bm{r}}$ and, as a consequence, on ray density $\sigma_{\bm{r}}$.
We further simplify the integral by changing the integration boundaries to $[0,1]$ and substituting $y_n = 0$.

Given the above derivation, we construct {\it the reparameterized Monte Carlo estimate} for the right-hand side integral in Eq.~\ref{eq:y_reparameterization_3}
\begin{equation}\label{eq:reparameterized_mc_color_approximation}
\hat{C}_{MC}^{R}(\bm{r}) := \frac{y_f}{k} \sum_{i=1}^k c_{\bm{r}}(F_{\bm{r}}^{-1}(y_f u_i)),
\end{equation}
 with $k$ i.i.d. $U[0, 1]$ samples $u_1, \dots, u_k$. It is easy to show that the estimate in Eq.~\ref{eq:reparameterized_mc_color_approximation} is an unbiased estimate of expected color in Eq.~\ref{eq:expected_color} and its gradient is an unbiased estimate of the gradient of the expected color $C(\bm{r})$. Additionally, we propose to replace the uniform samples $u_1,\dots,u_k$ with uniform independent samples within regular grid bins $v_i \sim U[\tfrac{i - 1}{k+1}, \tfrac{i}{k + 1}], i=1,\dots,k$. The latter samples yield a stratified variant of the estimate in Eq.~\ref{eq:reparameterized_mc_color_approximation} and, most of the time, lead to lower variance estimates (see Appendix~\ref{sec:toy_exp}).

In the above estimate, random samples $u_1, \dots, u_k$ do not depend on volume density $\sigma_{\bm{r}}$ or color $c_{\bm{r}}$. Essentially, for the reparameterized Monte Carlo estimate we generate samples from $p_{\bm{r}}(t)$ using inverse cumulative distribution function $F_{\bm{r}}^{-1}(y_f u)$. In what follows, we coin the term {\it reparameterized volume sampling (RVS)} for the sampling procedure. However, in practice, we cannot compute $F_{\bm{r}}$ analytically and can only query $\sigma_{\bm{r}}$ at certain ray points. Thus, in the following section, we introduce approximations of $F_{\bm{r}}$ and its inverse.

\subsection{Opacity Approximations}
\label{sec:opacity_approx}

\begin{figure}[t]
\centering
\includegraphics[width=1.0\linewidth]{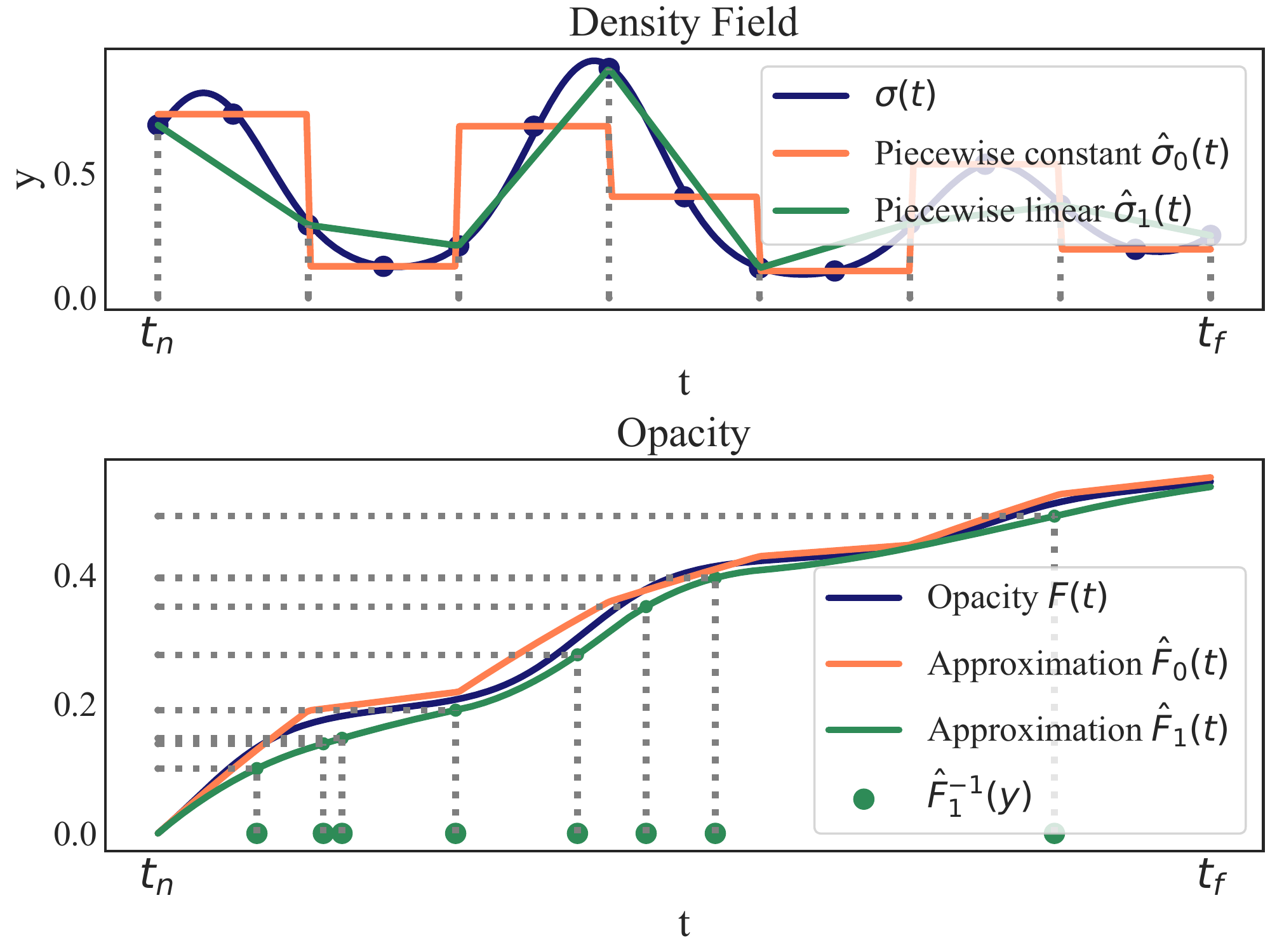}
\caption{Illustration of opacity inversion. On the left, we approximate density field $\sigma_{\bm{r}}$ with a piecewise constant and a piecewise linear approximation. On the right, we approximate opacity $F_{\bm{r}}(t)$ and compute~$F^{-1}_{\bm{r}}(y_f u)$ for $u \sim U[0, 1]$.}
\label{fig:spline_inversion}
\end{figure}

The expected radiance estimate in Eq.~\ref{eq:reparameterized_mc_color_approximation} relies on opacity $F_{\bm{r}}(t) = 1 - \exp \left(-\int_{t_n}^t \sigma_{\bm{r}}(s) \mathrm d s \right)$ and its inverse $F^{-1}_{\bm{r}}(y)$. We propose to approximate the opacity using a piecewise density field approximation. Fig.~\ref{fig:spline_inversion} illustrates the approximations and ray samples obtained through opacity inversion.
To construct the approximation, we take a grid $t_n = t_0 < t_1 < \dots < t_m = t_f$ and construct either a piecewise constant or a piecewise linear approximation. In the former case, we pick a point within each bin $t_{i} \leq \hat{t}_i \leq t_{i + 1}$ and approximate density with $\sigma_{\bm{r}}(\hat{t}_i)$ inside the corresponding bin. In the latter case, we compute $\sigma_{\bm{r}}$ in the grid points and interpolate the values between the grid points. Importantly, for a non-negative field these two approximations are also non-negative. Then we compute $\int_{t_n}^{t} \sigma_{\bm{r}}(s) \mathrm ds$ , which is as a sum of rectangular areas in the piecewise constant case
\begin{equation}
I_0(t) = \sum_{j=1}^{i} \sigma_{\bm{r}}(\hat{t}_j) (t_j - t_{j - 1}) + \sigma_{\bm{r}} (\hat{t}_i) (t - t_i).
\end{equation}

Analogously, the integral approximation $I_1(t)$ in the piecewise linear case is a sum of trapezoidal areas.

Given these approximations, we can approximate $y_f$ and $F_{\bm{r}}$ in Eq.~\ref{eq:reparameterized_mc_color_approximation}.
We generate samples on a ray based on inverse opacity $F^{-1}_{\bm{r}}(y)$ by solving the equation
\begin{equation}
y_f u = F_{\bm{r}}(t) = 1 - \exp\left( -\int_{t_n}^t \sigma_{\bm{r}}(s) \mathrm d s \right)
\end{equation}
for $t$, where $u \in [0, 1]$ is a random sample. We rewrite the equation as $- \log(1 - y_f u) = \int_{t_n}^t \sigma_{\bm{r}}(s) \mathrm d s$ and note that integral approximations $I_0(t)$ and $I_1(t)$ are monotonic piecewise linear and piecewise quadratic functions. We obtain the inverse by first finding the bin that contains the solution and then solving a linear or a quadratic equation. Crucially, the solution $t$ can be seen as a differentiable function of density field $\sigma_{\bm{r}}$ and we can back-propagate the gradients w.r.t. $\sigma_{\bm{r}}$ through $t$. We provide explicit formulae for $t$ for both approximations in Appendix~\ref{sec:inverse_explicit} and discuss the solutions crucial for the numerical stability in Appendix~\ref{sec:inverse_stability}. In Appendix~\ref{sec:sampling_pseudocode}, we provide the algorithm implementation and draw parallels with earlier work. Additionally, in Appendix~\ref{sec:inverse_implicit} we discuss an alternative approach to calculating inverse opacity and its gradients. We use piecewise linear approximations in Subsection~\ref{sec:nerf} and piecewise constant in Subsection~\ref{sec:dvgo}. 

\subsection{Application to Hierarchical Sampling}
\label{sec:hierarchical_rvs}

Finally, we propose to apply our RVS algorithm to the hierarchical sampling scheme originally proposed in NeRF. Here we do not change the final color approximation, utilizing the original one (Eq.~\ref{eq:grid_color_approximation}), but modify the way the coarse density network is trained. The method we introduce consists of two changes to the original scheme. Firstly, we replace sampling from piecewise constant PDF along the ray defined by weights $w_i$ (see Section~\ref{sec:nerf-recap}) with our RVS sampling algorithm that uses piecewise linear approximation of $\sigma_{\bm{r}}$ and generates samples from $p_r(t)$ using inverse CDF. Secondly, we remove the auxiliary reconstruction loss imposed on the coarse network. Instead, we propagate gradients through sampling. This way, we eliminate the need for auxiliary coarse network losses and train the network to solve the actual task of our interest: picking the best points for evaluation of the fine network. All components of the model are trained together end-to-end from scratch. In Subsection~\ref{sec:nerf}, we refer to the coarse network as the proposal network, since such naming better captures its purpose.

\section{RELATED WORK}\label{sec:related-work}

{\bf Monte Carlo estimates for integral approximations.}
In this work, we revisit the algorithm introduced to approximate the expected color in~\cite{max1995optical}. Currently, it is the default solution in multiple works on neural radiance fields. \cite{max1995optical} approximate density and radiance fields with a piecewise constant functions along a ray and compute Eq.~~\ref{eq:expected_color} as an approximation. Instead, we reparameterize Eq.~\ref{eq:expected_color} and construct Monte Carlo estimates for the integral. To compute the estimates in practice we use piecewise approximations only for the density field.
The cumulative density function (CDF) used in our estimates involves integrating the density field along a ray. \cite{lindell2021autoint} construct field anti-derivatives to accelerate inference. While they use the anti-derivatives to compute~\ref{eq:expected_color} on a grid with fewer knots, the anti-derivatives can be applied in our sampling method based on the inverse CDF without resorting to piecewise approximations.

In the past decade, integral reparameterizations have become a common practice in generative modeling~\citep{kingma2013auto, rezende2014stochastic} and approximate Bayesian inference~\citep{blundell2015weight,gal2016dropout,molchanov2017variational}. Similar to Equation~\ref{eq:expected_color}, objectives in these areas require optimizing expected values with respect to distribution parameters. We refer readers to~\cite{mohamed2020monte} for a systematic overview. Notably, in computer graphics, \cite{loubet2019reparameterizing} apply reparameterization to estimate gradients of path-traced images with respect to scene parameters.

{\bf NeRF acceleration through architecture and sparsity.} Since the original NeRF work~\citep{mildenhall2020nerf}, a number of approaches that aim to improve the efficiency of the model have been proposed. One family of methods tries to reduce the time required to evaluate the field. It includes a variety of architectures combining Fourier features~\citep{tancik2020fourier} and grid-based features~\citep{garbin2021fastnerf, sun2022direct, fridovich2022plenoxels, reiser2021kilonerf}. Besides grids, some works exploit space partitions based on Voronoi diagrams~\citep{rebain2021derf}, trees~\citep{hu2022efficientnerf, yu2021plenoctrees} and even hash tables~\citep{mueller2022instant}. These architectures generally trade-off inference speed for parameter count. TensorRF~\citep{chen2022tensorf} stores the grid tensors in a compressed format to achieve both high compression and fast performance. On top of that, skipping field queries for the empty parts of a scene additionally improves rendering time~\citep{levoy1990efficient}. Recent works (such as~\cite{hedman2021baking, fridovich2022plenoxels, liu2020neural, li2022nerfacc, sun2022direct, mueller2022instant}) manually exclude low-weight components in Eq~\ref{eq:grid_color_approximation} to speed up rendering during training and inference. Below, we show that our Monte Carlo algorithm is compatible with fast architectures and sparse density fields, achieving comparable speedups by using a few radiance evaluations.

{\bf Anti-aliased scene representations.} Mip-NeRF~\citep{barron2021mip}, Mip-NeRF 360~\citep{barron2022mip} and a more recent Zip-NeRF~\citep{barron2023zip} represent a line of work that modifies scene representations. Relevant to our research is the fact that these models employ modifications of the original hierarchical sampling scheme, where the coarse network parameterizes some density field. Mip-NeRF parameterizes the coarse and the fine fields by the same neural network that represents the scene at a continuously-valued scale. Mip-NeRF 360 and Zip-NeRF use a separate model for proposal density,  but train it to mimic the fine density rather than independently reconstructing the image. This means that our method for training the proposal density field can be potentially used to improve the performance of these models and simplify the training algorithm.

{\bf Algorithms for picking ray points.}
\cite{mildenhall2020nerf} employs a hierarchical scheme to generate ray points using an auxiliary density and color fields. Since then, a number of other methods for picking ray points, which focus on real-time rendering and aim to improve the efficiency of NeRF, have been proposed. DoNeRF~\citep{neff2021donerf} uses a designated depth oracle network supervised with ground truth depth maps. TermiNeRF~\citep{piala2021terminerf} foregoes the depth supervision by distilling the sampling network from a pre-trained NeRF model. NeRF-ID~\citep{arandjelovic2021nerf} adds a separate differentiable proposer neural network to the original NeRF model that maps outputs of the coarse network into a new set of samples. The model is trained in a two-stage procedure together with NeRF. The authors of NeuSample~\citep{fang2021neusample} use a sample field that directly transforms rays into point coordinates. The sample field can be further fine-tuned for rendering with a smaller number of samples. AdaNeRF~\citep{kurz2022adanerf} proposes to use a sampling and a shading network. Samples from the sampling network are processed by the shading network that tries to predict the importance of samples and cull the insignificant ones. One of the key merits of our approach in comparison to these works is its simplicity. We simplify the original NeRF training procedure, while other works only build upon it, adding new components, training stages, constraints, or losses. Moreover, the absence of reliance on additional neural network components (not responsible for density or radiance) for sampling makes our approach better suited for fast NeRF architectures. Finally, our approach is suitable for end-to-end training of NeRF models from scratch, whereas the works mentioned above use pre-trained NeRF models or multiple training stages. 

\section{EXPERIMENTS}

\label{sec:experiments}
\subsection{End-to-end Differentiable Hierarchical Sampling}
\label{sec:nerf}

In this section, we evaluate the proposed approach to hierarchical volume sampling (see Subsection~\ref{sec:hierarchical_rvs}).

\begin{figure*}[t]
\centering
\includegraphics[width=1.0\linewidth]{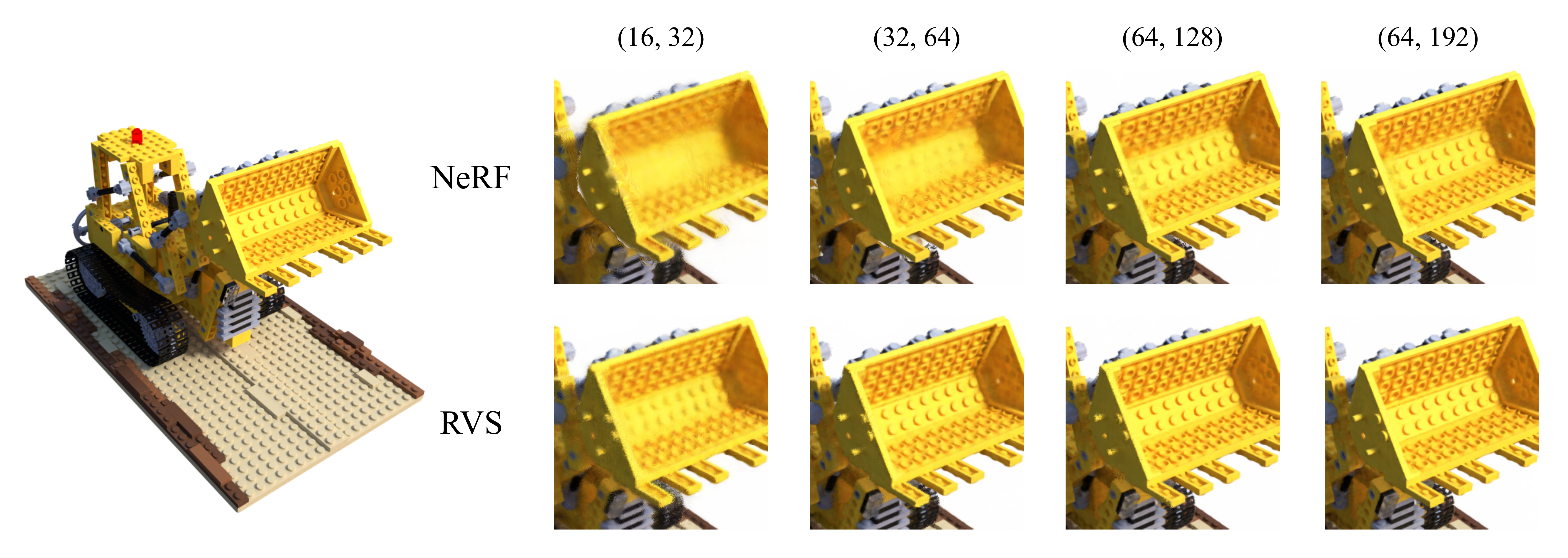}
\caption{Comparison between renderings of test-set view on the Lego scene (Blender). Rows correspond to different $(N_p, N_f)$ configurations. NeRF baseline has difficulties in reconstructing fine details for $(16, 32)$ and $(32, 64)$ configurations, and some parts remain blurry even in $(64, 128)$ and $(64, 192)$ configurations, while our method already produces realistic reconstruction for $(32, 64)$ configuration.}
\label{fig:nerf_lego} 
\end{figure*}

{\bf Experimental setup.} We do the comparison by fixing some training setup and training two models from scratch: one NeRF model is trained using the procedure proposed in~\cite{mildenhall2020nerf} (further denoted as NeRF in the results), and the other one is trained using our modification described in Subsection~\ref{sec:hierarchical_rvs} (further denoted as RVS). For both models, the final color approximation is computed as in Eq.~\ref{eq:grid_color_approximation}, so the difference only appears in the proposal component. 

We train all models for 500k iterations using the same hyperparameters as in the original paper with minor differences. We replace ReLU density output activation with Softplus. The other difference is that we use a smaller learning rate for the proposal density network in our method (start with $5 \times 10^{-5}$ and decay to $5 \times 10^{-6}$) but the same for the fine network (start with $5 \times 10^{-4}$ and decay to $5 \times 10^{-5}$). This is done since we observed that decreasing the proposal network learning rate improves stability of our method. We run the default NeRF training algorithm in our experiments with the same learning rates for proposal and fine networks. Further in the ablation study we show that decreasing the proposal network learning rate only degrades the performance of the base algorithm. We use PyTorch implementation of NeRF~\citep{lin2020nerfpytorch} in our experiments.
\begin{table}
\centering
\small
\caption{Comparison on the Lego scene of Blender dataset between NeRF training algorithm and our modification depending on the number of proposal and fine network evaluations per ray $(N_p, N_f)$. The $(64, 192)$ configuration is the one originally used in NeRF. The training time column depicts relative training time on a single NVIDIA A100 GPU (1.0 being 12 hours).}
\label{tab:nerf_lego} 
\resizebox{0.485\textwidth}{!}{\begin{tabular}{l|cc|c|ccc}
 & \multicolumn{2}{c|}{Evals.} & Train & PSNR $\uparrow$ & SSIM $\uparrow$ & LPIPS $\downarrow$ \\
& $N_p$ & $N_f$  & time & \\
\hline
NeRF     & 16     & 32  & 0.39  & 27.09     & 0.913  & 0.121 \\
RVS      & 16     & 32 & 0.37   & \textbf{29.18}     & \textbf{0.928}  & \textbf{0.112} \\
\hline
NeRF      & 32     & 64  & 0.52  & 30.11     & 0.947  & 0.070 \\
RVS      & 32     & 64 & 0.48   & \textbf{31.89}     & \textbf{0.955}  & \textbf{0.066} \\
\hline
NeRF      & 64     & 128  & 0.79  & 32.14    & 0.958  & 0.053 \\
RVS      & 64     & 128 & 0.76   & \textbf{32.80}    & \textbf{0.963}  & \textbf{0.051} \\
\hline
NeRF      & 64     & 192  & 1.0  & 32.69    & 0.962  & \textbf{0.048} \\
RVS      & 64     & 192  & 0.98  & \textbf{33.03}    & \textbf{0.964}  & \textbf{0.047}\\
\end{tabular}}
\end{table}

{\bf Comparative evaluation.} We start the comparison on the Lego scene of the synthetic Blender dataset~\citep{mildenhall2020nerf} for different $(N_p, N_f)$ configurations that correspond to the number of proposal and fine network evaluations. Note that this $N_p$, $N_f$ notation does not directly correspond to the $N_c$, $N_f$ notation used in Section~\ref{sec:nerf-recap} since the original NeRF model evaluates the fine network in $N_c$ + $N_f$ points. For more details on training configurations and options for picking points for fine network evaluation, see Appendix \ref{sec:hierarchical_details}. The results are presented in Table~\ref{tab:nerf_lego}. Our method outperforms the baseline across all configurations and all metrics, with the only exception of LPIPS in $(64, 192)$ configuration, where it showed similar performance. We observe that the improvement is more significant for smaller $(N_p, N_f)$. Our method also has a minor speedup over the baseline due to the fact that the former does not use the radiance component of the proposal network. Fig.~\ref{fig:nerf_lego} in visualizes test-set view renderings of models trained by two methods.

\begin{figure}[t!]
\includegraphics[width=1.0\linewidth]{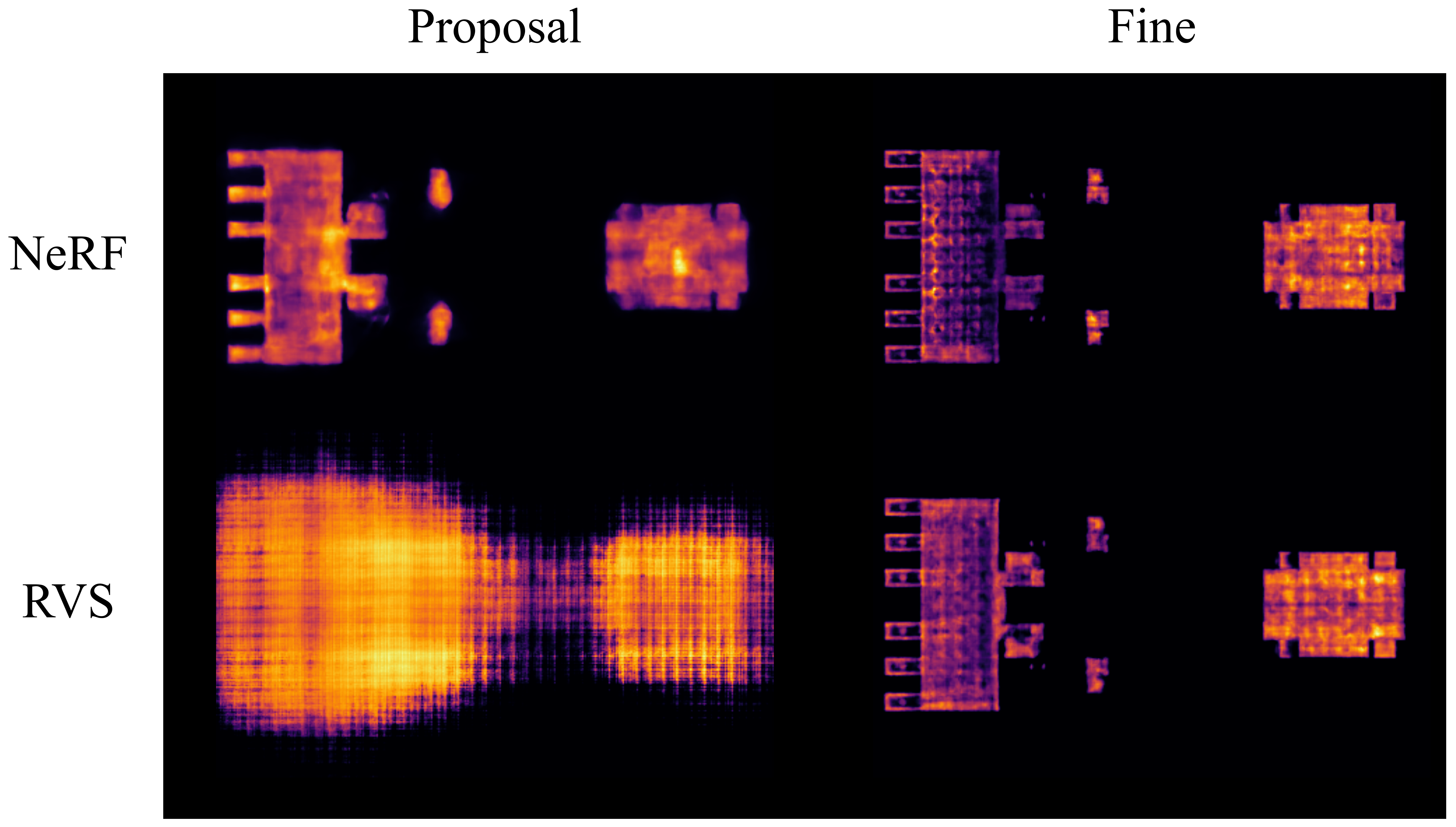}
\caption{Visualizations across 2D slice of proposal and fine densities learnt on the Lego scene in $(32, 64)$ configuration. Brighter pixels correspond to larger density values. 
}
\label{fig:nerf_prop_viz}
\end{figure}

We also visualize proposal and fine densities learned by two algorithms in Fig.~\ref{fig:nerf_prop_viz}. Figures are constructed by fixing some value of $z$ coordinate and calculating the density on $(x, y)$-plane. While fine density visualizations look similar, proposal densities turn out very different. This happens due to the fact that the original algorithm trains the proposal network to reconstruct the scene (but using a smaller number of points for color estimation), while our algorithm trains this network to sample points for fine network evaluation that would lead to a better overall reconstruction. Since RVS shows better reconstruction quality, and its proposal densities are significantly different from the NeRF ones, one can argue that a proposal density that tries to mimic the fine density is not the optimal choice.

\begin{table}[t!]
\small
\centering
\caption{Comparison with NeRF on Blender and LLFF datasets in $(N_p = 32, N_f = 64)$ configuration.}
\begin{tabular}{l|ccc}
& \multicolumn{3}{c}{Blender Dataset}   \\
& PSNR $\uparrow$ & SSIM $\uparrow$ & LPIPS $\downarrow$  \\
\hline
NeRF & 29.49     & 0.934    & 0.085      \\
RVS & \textbf{30.26}     & \textbf{0.939}    & \textbf{0.082}     \\

& \multicolumn{3}{c}{LLFF Dataset}   \\
& PSNR $\uparrow$ & SSIM $\uparrow$ & LPIPS $\downarrow$  \\
\hline
NeRF  & 26.01     & 0.796    & 0.273      \\
RVS  & \textbf{26.24}     & \textbf{0.799}    & \textbf{0.270}     \\
\end{tabular}
\label{tab:nerf_blender_llff} 
\end{table}

Next, we evaluate our method on all scenes from Blender, as well as the LLFF~\citep{mildenhall2019local} dataset containing real scenes. Table~\ref{tab:nerf_blender_llff} depicts the results averaged across all scenes. The results for individual scenes can be found in Appendix~\ref{appendix:per-scene-results}.  
Our approach shows improvement over the baseline on all scenes of Blender dataset. While the improvement is less pronounced on LLFF dataset, it is still present across all metrics on average. Fig.~\ref{fig:nerf_trex} in Appendix~\ref{appendix:visualizations} visually depicts the quality of reconstruction on T-Rex scene.

{\bf Unbounded scenes.} In addition, we evaluate our approach with NeRF++ \citep{zhang2020nerf++} modification designed for unbounded scenes. NeRF++ utilizes the same hierarchical scheme as the original NeRF, so we ran the same setup as previously: one model is trained using the original procedure, and for the other one we replace the sampling algorithm with RVS and propagate gradients through sampling instead of using a separate reconstruction loss for the proposal network. We did not modify any hyperparameters in comparison to \cite{zhang2020nerf++} apart from using a smaller $(N_p, N_f)$ configuration and a smaller proposal learning rate in our approach, the same as in the previous experiments. We run the comparison on LF~\citep{yucer2016efficient} and T\&T~\citep{knapitsch2017tanks} datasets containing unbounded real scenes. The results are presented in Table~\ref{tab:nerfplusplus}. Our approach also shows improvement over NeRF++ across all metrics on both datasets. Table~\ref{tab:nerfpp-lf-all} and Table~\ref{tab:nerfpp-tt-all} in Appendix~\ref{appendix:per-scene-results} present the results for individual scenes.
\begin{table}[t!]
\small
\centering
\caption{Comparison with NeRF++ on LF and T\&T datasets in $(N_p = 32, N_f = 64)$ configuration.}
\begin{tabular}{l|ccc}
& \multicolumn{3}{c}{LF Dataset}   \\
& PSNR $\uparrow$ & SSIM $\uparrow$ & LPIPS $\downarrow$  \\
\hline
NeRF++ & 23.99     & 0.784    & 0.287      \\
RVS & \textbf{24.63}     & \textbf{0.812}    & \textbf{0.253}     \\

& \multicolumn{3}{c}{T\&T Dataset}   \\
& PSNR $\uparrow$ & SSIM $\uparrow$ & LPIPS $\downarrow$  \\
\hline
NeRF++  & 19.21     & 0.612    & 0.493      \\
RVS  & \textbf{19.62}     & \textbf{0.622}    & \textbf{0.472}     \\
\end{tabular}
\label{tab:nerfplusplus} 
\end{table}
\begin{table}[t!]
\centering
\caption{Ablation study with NeRF on Blender dataset in $(N_p = 32, N_f = 64)$ configuration.}
\resizebox{0.485\textwidth}{!}{\begin{tabular}{l|ccc}
& \multicolumn{3}{c}{Blender Dataset}   \\
& PSNR $\uparrow$ & SSIM $\uparrow$ & LPIPS $\downarrow$  \\
\hline
aux. loss (base NeRF) & 29.49     & 0.934    & 0.085      \\
aux. loss (smaller prop. lr) & 29.23     & 0.932    & 0.090      \\
aux. loss (union of points) & 29.06     & 0.929    & 0.096      \\
aux. loss (our sampling) & 29.70     & 0.936    & \textbf{0.082}      \\
end-to-end (NeRF sampling) & 29.07     & 0.929    & 0.095      \\
end-to-end (our sampling) & \textbf{30.26}     & \textbf{0.939}    & \textbf{0.082}     \\
\end{tabular}}
\label{tab:nerf_ablation} 
\end{table}

{\bf Ablation study.}  Finally, we ablate the influence of some components of our approach on the results. 
Table~\ref{tab:nerf_ablation} presents the ablation study. Firstly, we run NeRF baseline with a decreased proposal network learning rate, thus fully matching all training hyperparameters with our method. This only reduces all metrics. Next, we run our approach that trains the proposal network end-to-end, but replace our algorithm that draws samples from the proposal distribution with the sampling algorithm originally proposed in NeRF. Even though the original work does not propagate gradients through sampling, the algorithm is still end-to-end differentiable, thus the setup is plausible. It also produces results that fall behind the baseline NeRF. This shows that while both algorithms are differentiable, ours is better suited for end-to-end optimization. We discuss some possible reasons behind these results and the differences between the two algorithms in Appendix~\ref{sec:sampling_pseudocode}. After that, we run the original NeRF approach that uses an auxiliary loss for proposal network training, but replace the sampling algorithm with ours (without propagating gradients through sampling). This variant performs better than the baseline, but still falls behind end-to-end training with RVS. Finally, we run the baseline with a different strategy for picking fine points (see Appendix \ref{sec:hierarchical_details} for a detailed discussion), which also leads to degradation of the baseline performance.

\subsection{Scene Reconstruction with Monte Carlo Estimates}
\label{sec:dvgo}
In this section, we evaluate the proposed Monte Carlo radiance estimate (see Subsection~\ref{sec:mc_est} and Eq.~\ref{eq:reparameterized_mc_color_approximation}) as a part of the rendering algorithm for scene reconstruction. Given an accurate density field approximation, our color estimate is unbiased for any given number of samples $k$. Therefore, our estimate is especially suitable for architectures that can evaluate the density field faster than the radiance field. As an example, we pick a voxel-based radiance field model DVGO~\citep{sun2022direct} that parametrizes density field as a voxel grid and relies on a combination of voxel grid and a view-dependent neural network to parameterize radiance. In Appendix~\ref{appendix:computational_efficiency}, we evaluate the inference time of the model to illustrate the benefits of our approach. In our experiments, we take the default model parameters and only replace the rendering algorithm. 
\begin{figure}[t!]
\centering
\includegraphics[width=\linewidth]{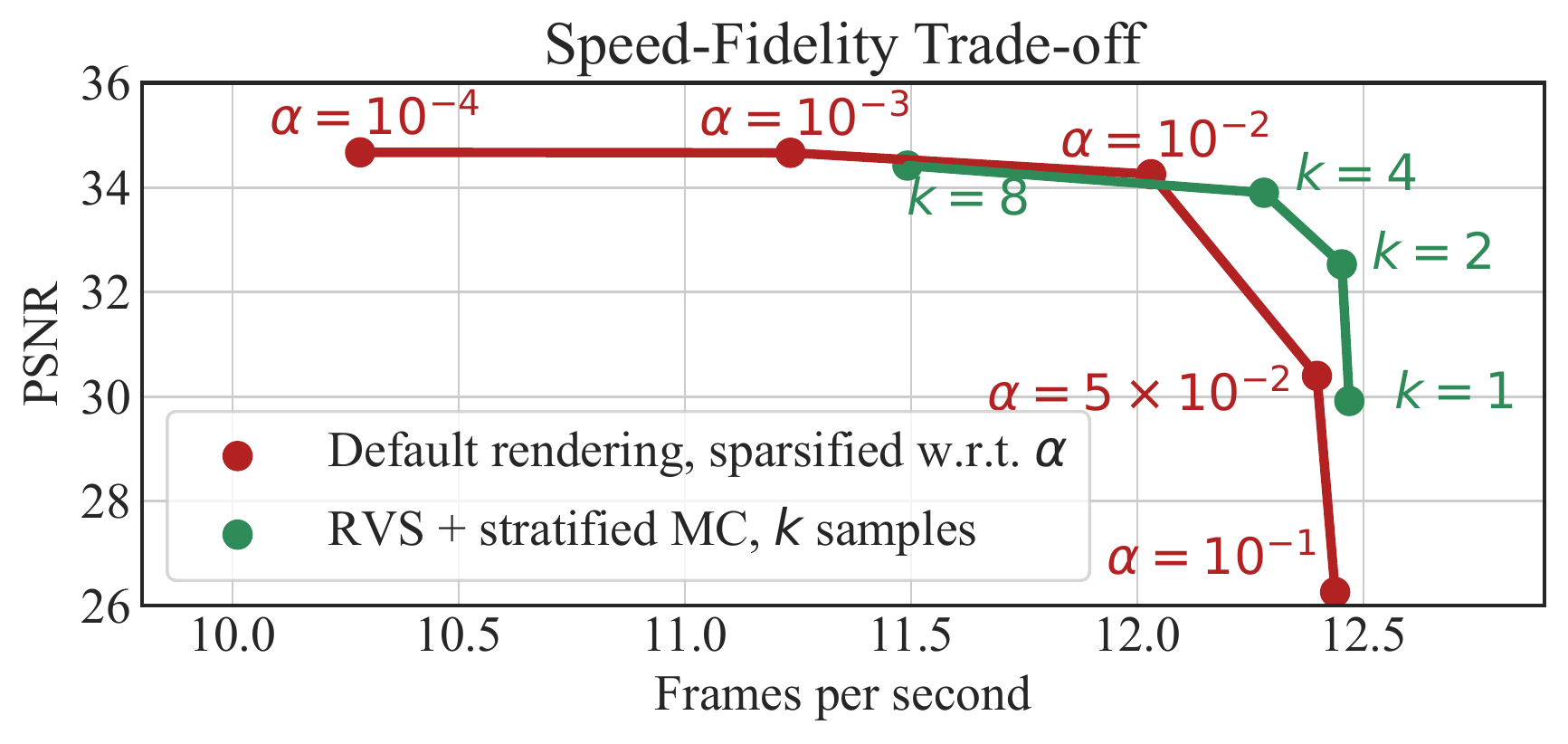}
\caption{Test-time rendering quality as a function of rendering speed. Given a scene representation pre-trained with the standard rendering algorithm, we evaluate rendering algorithms at various configurations.}
\label{fig:rendering_ablation}
\end{figure}

{\bf Experimental setup \& comparative evaluation.} In Table~\ref{tab:dvgo_it_s}, we report iterations per second and peak memory consumption on the Lego scene of Blender dataset during training. One of the primary goals of DVGO is to reduce the training time of a scene model. To achieve that goal, the authors mask components in the sum in Eq.~\ref{eq:grid_color_approximation} that have weights below a certain threshold $\alpha$. We achieve a similar effect without thresholding: $k=8$ samples yield a comparable $\times 6$ speedup and with fewer samples, we reduce iteration time even further. However, the speed-up in our estimate comes at the cost of additional estimate variance. Next, we evaluate the effect of additional variance on the rendering fidelity. In Fig.~\ref{fig:rendering_ablation}, we report test PSNR on a fixed pre-trained Lego scene representation for both rendering algorithms, comparing them at the \textit{inference stage}. Our algorithm achieves the same average PSNR with $k=8$ samples and outperforms the default rendering algorithm in case of a higher sparsity threshold $\alpha$. 

\begin{table}[t!]
\small
\centering
\caption{Training iteration times and peak memory consumption for different color approximations.}
\begin{tabular}{l|cc}
DVGO Renderer                  & Speed & Memory \\
\hline
Default w/o sparisty  & 24 it/s                 & 9 GB             \\
Default w/ sparsity & 160 it/s                & 5 GB             \\
MC + RVS, $k=128$ & 20 it/s                 & 8 GB             \\
MC + RVS, $k=8$   & 170 it/s                & 5 GB             \\
MC + RVS, $k=1$   & 260 it/s                & 5 GB            
\end{tabular}
\label{tab:dvgo_it_s}
\end{table}

\begin{table}[t!]
\centering
\small
\caption{Scene reconstruction results on Blender for training with Monte Carlo estimates.}
\resizebox{0.48\textwidth}{!}{\begin{tabular}{l|cccc}
DVGO Renderer & \begin{tabular}{@{}c@{}}Train\\ time \\ (min)\end{tabular}   & PSNR$\uparrow$  & SSIM$\uparrow$  & LPIPS$\downarrow$ \\
\hline
Default w/ sparsity                      & 2:48 & {\bf 31.90} & {\bf 0.956} & {\bf 0.054} \\
 MC + RVS, $k=4$                & 2:24 &       31.19 &       0.951 &       0.059 \\
 MC + RVS, $k=8$                & 2:00 &       31.13 &       0.951 &       0.059 \\
 MC + RVS, \tiny{adaptive} $k$ & 2:54 &       31.44 &       0.953 &       0.056
\end{tabular}}
\label{tab:dvgo} 
\end{table}

In Table~\ref{tab:dvgo}, we report the performance of models trained with various color estimates (see Appendix~\ref{appendix:per-scene-results} for per-scene results), comparing them at the \textit{training stage}.
We train a model with $k=4$ samples using $\times \tfrac{3}{2}$ more training steps than with $k=8$ samples aiming to achieve similar training times.
Additionally, we consider a model that chooses the number of samples $k$ on each ray adaptively between $4$ and $48$ based on the number of grid points with high density.
The adaptive number of samples allows to reduce estimate variance without a drastic increase in training time. We evaluate our models with $k=64$ samples to mitigate the effect of variance on evaluation.

As Table~\ref{tab:dvgo} indicates, the training algorithm
with our color approximation fails to outperform the base algorithm
in terms of reconstruction quality; however, it allows for the faster
training of the model. 

{\bf Ablation study.} We ablate the proposed algorithm on the Lego scene to gain further insights into the difference in reconstruction fidelity when it is used for training. Specifically, we ablate parameters affecting optimization aiming to match the reconstruction quality of the default algorithm. The results are given in Table~\ref{tab:optimization_ablation}. The increase in training steps or the number of samples improves the performance but does not lead to matching results. Increased spline density improves both our model and the baseline to the same extent. We also noticed that the standard objective estimates the expected loss $\mathbb E L_2(\hat{C}, C_{gt})$ rather than the loss at expected radiance $L_2(\mathbb E \hat{C}, C_{gt})$, which leads to an additional bias towards low-variance densities. The estimate $(\hat{C}_1 - C_{gt}) (\hat{C}_2 - C_{gt})$ with two i.i.d. color estimates is an unbiased estimate of the latter and allows reconstructing non-degenerate density fields, but the estimate has little effect in case of degenerate densities omnipresent in 3D scenes. Finally, with a denser ray grid our algorithm surpasses the baseline PSNR at the cost of increased training time. For qualitative comparison, in Appendix~\ref{appendix:visualizations} we visualize the differences in reconstructed models and in renders reported in Fig.~\ref{fig:rendering_ablation}.

\begin{table}[t!]
\centering
\caption{Ablation study for training with Monte Carlo estimates on the Lego scene.}
\begin{tabular}{ll}
Ablated Feature                                & PSNR    \\
\hline
MC + RVS with adaptive $k$                         & \ $33.85$ \\
\hline
Training steps $(\times 7)$                          & $+0.28$ \\
Monte Carlo samples $(\times 2 \text{\ on avg.})$      & $+0.15$ \\
Dense grid on rays $(\times 2)$                      & $+0.29$ \\
Unbiased loss $L_2(\mathbb E \hat{C}, C_{gt})$       & $+0.25$ \\
Density grid resolution $(160^3 \rightarrow 256^3)$  & $+0.63$ \\
Radiance grid resolution $(160^3 \rightarrow 256^3)$ & $+0.08$ \\
$\sigma$ + $c$ resolution $(160^3 \rightarrow 256^3)$& {\bf+0.86} \\
\hline
DVGO                                                 & \ $34.64$
\end{tabular}
\label{tab:optimization_ablation}
\end{table}

To summarize, training with the reparameterized Monte Carlo estimate currently does not fully match the fidelity of the standard approach. At the same time, Monte Carlo radiance estimates provide a straightforward mechanism to control both training and inference speed.

\section{CONCLUSION}
The core of our contribution is an end-to-end differentiable ray point sampling algorithm.
We utilize it to construct an alternative rendering algorithm based on Monte Carlo, which provides an explicit mechanism to control rendering time during the inference and training stages.
While it is able to outperform the standard rendering algorithm at the inference stage given a pre-trained model, it achieves lower reconstruction quality when used during training, which suggests areas for future research.
At the same time, we show that the proposed sampling algorithm improves scene reconstruction in hierarchical models and simplifies the training approach by disposing of auxiliary losses.

\section*{Acknowledgements}
The study was carried out within the strategic project "Digital Transformation: Technologies, Effects and Performance", part of the HSE University "Priority 2030" Development Programme. The results on hierarchical sampling from Section~\ref{sec:nerf} were obtained by Kirill Struminsky with the support of the grant for research centers in the field of AI provided by the Analytical Center for the Government of the Russian Federation (ACRF) in accordance with the agreement on the provision of subsidies (identifier of the agreement 000000D730321P5Q0002) and the agreement with HSE University No. 70-2021-00139. This research was supported in part through computational resources of HPC facilities at HSE University~\citep{kostenetskiy2021hpc}.

\bibliographystyle{apalike}
\bibliography{ref}
\section*{Checklist}

 \begin{enumerate}

 \item For all models and algorithms presented, check if you include:
 \begin{enumerate}
   \item A clear description of the mathematical setting, assumptions, algorithm, and/or model. [Yes, see Section~\ref{sec:nerf-recap} and Section~\ref{sec:theory}]
   \item An analysis of the properties and complexity (time, space, sample size) of any algorithm. [Yes, see Appendix~\ref{sec:toy_exp} and Appendix~\ref{appendix:computational_efficiency} for numerical analysis]
   \item (Optional) Anonymized source code, with specification of all dependencies, including external libraries. [Yes, see Section~\ref{gen_inst}]
 \end{enumerate}

 \item For any theoretical claim, check if you include:
 \begin{enumerate}
   \item Statements of the full set of assumptions of all theoretical results. [Not Applicable]
   \item Complete proofs of all theoretical results. [Yes, see Section~\ref{sec:theory} and Appendix~\ref{sec:inverse_explicit}] 
   \item Clear explanations of any assumptions. [Yes, see Section~\ref{sec:theory}] 
 \end{enumerate}

 \item For all figures and tables that present empirical results, check if you include:
 \begin{enumerate}
   \item The code, data, and instructions needed to reproduce the main experimental results (either in the supplemental material or as a URL). [Yes, see Section~\ref{gen_inst} and Section~\ref{sec:experiments}] 
   \item All the training details (e.g., data splits, hyperparameters, how they were chosen). [Yes, see Section~\ref{sec:experiments}]
         \item A clear definition of the specific measure or statistics and error bars (e.g., with respect to the random seed after running experiments multiple times). [No. We use well-established measures to report the performance of the proposed algorithm. We report the measures across a variety of datasets and a range of hyperparameters. We do not include results averaged across multiple runs as randon reinitialization has almost no effect on the resulting performance.]
         \item A description of the computing infrastructure used. (e.g., type of GPUs, internal cluster, or cloud provider). [Yes, see Section~\ref{sec:experiments}]
 \end{enumerate}

 \item If you are using existing assets (e.g., code, data, models) or curating/releasing new assets, check if you include:
 \begin{enumerate}
   \item Citations of the creator If your work uses existing assets. [Yes]
   \item The license information of the assets, if applicable. [Not Applicable]
   \item New assets either in the supplemental material or as a URL, if applicable. [Not Applicable]
   \item Information about consent from data providers/curators. [Not Applicable]
   \item Discussion of sensible content if applicable, e.g., personally identifiable information or offensive content. [Not Applicable]
 \end{enumerate}

 \item If you used crowdsourcing or conducted research with human subjects, check if you include:
 \begin{enumerate}
   \item The full text of instructions given to participants and screenshots. [Not Applicable]
   \item Descriptions of potential participant risks, with links to Institutional Review Board (IRB) approvals if applicable. [Not Applicable]
   \item The estimated hourly wage paid to participants and the total amount spent on participant compensation. [Not Applicable]
 \end{enumerate}

 \end{enumerate}

\newpage
\appendix
\onecolumn
\section{INVERSE OPACITY CALCULATION}

\subsection{Inverse Functions for Density Integrals}
\label{sec:inverse_explicit}
In this section, we derive explicit formulae for the density integral inverse used in inverse opacity.
\subsubsection{Piecewise Constant Approximation Inverse}
We start with a formula for the integral
\begin{equation}
I_0(t) = \sum_{j=1}^{i} \sigma_{\bm{r}}(\hat{t}_j) (t_j - t_{j - 1}) + \sigma_{\bm{r}} (\hat{t}_i) (t - t_i)
\end{equation}
and solve for $t$ equation
\begin{equation}
y = I_0(t).
\end{equation}
The equation above is a linear equation with solution
\begin{equation}
t = t_i + \frac{y - \sum_{j=1}^{i} \sigma_{\bm{r}}(\hat{t}_j) (t_j - t_{j - 1})}{\sigma_{\bm{r}} (\hat{t}_i)}.
\end{equation}
In our implementation we add small $\epsilon$ to the denominator to improve stability when $\sigma_{\bm{r}} (\hat{t}_i) \approx 0$.

\subsubsection{Piecewise Linear Approximation Inverse}
The piecewise linear density approximation yield a piecewise quadratic function
\begin{equation}
I_1(t) = \sum_{j=1}^{i} \frac{\sigma_{\bm{r}}(t_j) + \sigma_{\bm{r}}(t_{j-1})}{2} (t_j - t_{j - 1}) + \frac{(\sigma_{\bm{r}}(t_i) + \bar{\sigma}_{\bm{r}}(t))}{2} (t - t_i),
\end{equation}
where $\bar{\sigma}_{\bm{r}}(t) = \sigma_{\bm{r}}(t_i) \frac{t_{i+1} - t}{t_{i + 1} - t_i} + \sigma_{\bm{r}}(t_{i+1}) \frac{t - t_i}{t_{i + 1} - t_i}$ is the interpolated density at $t$. Again, we solve
\begin{equation}
y = I_1(t)
\end{equation}
for $t$. We change the variable to $\Delta t := t - t_i$ and note that terms $a$ and $c$ in quadratic equation 
\begin{equation}
0 = a \Delta t ^2 + b \Delta t + c
\end{equation}
will be
\begin{align}
a &= \frac{\sigma_{\bm{r}}(t_{i + 1}) - \sigma_{\bm{r}}(t_i)}{2} \\
c &= \left( \sum_{j=1}^{i} \frac{\sigma_{\bm{r}}(t_j) + \sigma_{\bm{r}}(t_{j-1})}{2} (t_j - t_{j - 1}) - y \right) \times (t_{i + 1} - t_i)
\end{align}
and with a few algebraic manipulations we find the linear term
\begin{equation}
b = \sigma_{\bm{r}}(t_i) \times (t_{i+1} - t_i).
\end{equation}
Since our integral monotonically increases, we can deduce that the root $\Delta t$ must be
\begin{equation}
\Delta t = \tfrac{-b + \sqrt{b^2 - 4ac}}{2a}.
\end{equation}
However, this root is computationally unstable when $a \approx 0$. The standard trick is to rewrite the root as
\begin{equation}
\Delta t = \tfrac{2c}{b + \sqrt{b^2 - 4ac}}.
\end{equation}
For computational stability, we add small $\epsilon$ to the square root argument. See the supplementary notebook for details.

\subsection{Numerical Stability in Inverse Opacity}
\label{sec:inverse_stability}

Inverse opacity input $y$ is a combination of a uniform sample $u$ and ray opacity $y_f = 1 - \exp \left(- \int_{t_n}^{t_f} \sigma_{\bm{r}}(s) \mathrm d s \right)$:
\begin{equation}
y = - \log(1 - y_f u).
\end{equation}
The expression above is a combination of a logarithm and exponent. We rewrite it to replace with more reliable $\operatorname{logsumexp}$ operator:
\begin{equation}
y = - \log \left( \exp(\log(1 - u)) + \exp( \log u - \int_{t_n}^{t_f} \sigma_{\bm{r}}(s) \mathrm d s ) \right).
\end{equation}
In practice, for opaque rays $\int_{t_n}^{t_f} \sigma_{\bm{r}} (s) \mathrm d s \approx 0$ implementation of $\operatorname{logsumexp}$ becomes computationally unstable. In this case, we replace $y$ with a first order approximation $u \cdot \int_{t_n}^{t_f} \sigma_{\bm{r}} (s) \mathrm d s$.

\subsection{Parallels with Prior Work and Algorithm Implementation}
\label{sec:sampling_pseudocode}
Original NeRF architecture uses inverse transform sampling to generate a grid for a fine network. They define a distribution based on Eq.~\ref{eq:nerf_weights} with piecewise constant density. In turn, we do inverse transform sampling from a distribution induced by a piecewise interpolation of $\sigma_{\bm r}$ in Eq.~\ref{eq:density_field_dist}. The two approximation approaches yield distinct sampling algorithms. In Listing~\ref{pseudocode}, we provide a numpy implementation of the two algorithms to highlight the difference. We rewrite NeRF sampling algorithm in an equivalent simplified form to facilitate the comparison. For simplicity, we provide implementation of RVS only for a piecewise constant approximation of $\sigma$. The inversion algorithm described in Appendix~\ref{sec:inverse_explicit} is hidden under the hood of {\it np.interp()}.

The interpolation scheme in our algorithm can be seen as linear interpolation of density field rather than probabilities. As a result, samples follow the Beer-Lambert law within each bin. The law describes light absorption in a homogeneous medium. In contrast, the hierarchical sampling scheme in NeRF interpolates exponentiated densities. We speculate that the numerical instability of the exponential function makes the latter algorithm less suitable for end-to-end optimization (as observed in Table~\ref{tab:nerf_ablation}).

\vspace{0.3cm}

\begin{lstlisting}[language=Python, caption=Numpy implementation of the inverse transform sampling procedure proposed in this work and the inverse transform sampling proposed in NeRF. The implementation assumes a single ray for brevity., label=pseudocode]
import numpy as np

def inverse_cdf(u, sigmas, ts, sampling_mode):
    """ Get inverse CDF sample

    Arguments:
    u - array of uniform random variables
    sigmas - array of density values on a grid
    ts - array of grid knots
    """
    # Compute $\int_{t_0}^{t_i} \sigma(s) ds$
    bin_integrals = sigmas * np.diff(ts)
    prefix_integrals = np.cumsum(bin_integrals)
    prefix_integrals = np.concatenate([np.zeros(1), prefix_integrals])
    # Get inverse CDF argument
    rhs = -np.expm1(-bin_integrals.sum()) * u
    if sampling_mode == 'rvs':  # interpolate $\int \sigma(s) ds$
        return np.interp(-np.log1p(-rhs), prefix_integrals, ts)
    elif sampling_mode == 'nerf':  # interpolate CDF
        return np.interp(rhs, -np.expm1(-prefix_integrals), ts)
\end{lstlisting}

\subsection{Implicit Inverse Opacity Gradients}
\label{sec:inverse_implicit}
To compute the estimates in Eq.~\ref{eq:reparameterized_mc_color_approximation}, we need to compute the inverse opacity $F_{\bm{r}}^{-1}(y)$ along with its gradient. 
In the main paper, we invert opacity explicitly with a differentiable algorithm. Alternatively, we could invert $F_{\bm{r}}(t) = 1 - \exp{\left( -\int_{t_n}^t \sigma_{\bm{r}}(s) \mathrm{d} s \right)}$ with binary search. This approach can be used in situations when the formula for inverse opacity cannot be explicitly derived.

Opacity $F_{\bm{r}}(t)$ is a monotonic function and for $y \in (y_n, y_f) = (F_{\bm{r}}(t_n), F_{\bm{r}}(t_f))$ the inverse lies in $(t_n, t_f)$. To compute $F_{\bm{r}}^{-1}(y)$, we start with boundaries $t_l = t_n$ and $t_r = t_f$ and gradually decrease the gap between the boundaries based on the comparison of $F_{\bm{r}}(\tfrac{t_l + t_r}{2})$ with $y$. Importantly, such procedure is easy to parallelize across multiple inputs and multiple rays.

However, we cannot back-propagate through the binary search iterations and need a workaround to compute the gradient $\tfrac{\partial t}{\partial \theta}$ of $t(\theta) = F_{\bm{r}}^{-1}(y, \theta)$. To do this, we follow~\cite{figurnov2018implicit} and compute differentials of the right and the left hand side of equation $y(\theta) = F_{\bm{r}}(t, \theta)$
\begin{equation}\label{eq:differential}
\frac{\partial y}{\partial \theta} \mathrm d \theta = \frac{\partial F_{\bm{r}}}{\partial t} \frac{\partial t}{\partial \theta} \mathrm d \theta + \frac{\partial F_{\bm{r}}}{\partial \theta} \mathrm d \theta.
\end{equation}
By the definition of $F_{\bm{r}}(t, \theta)$ we have
\begin{align}
\tfrac{\partial F_{\bm{r}}}{\partial t} &= (1 - F_{\bm{r}}(t, \theta)) \sigma_{\bm{r}} (t, \theta), \label{eq:dfdt} \\
\tfrac{\partial F_{\bm{r}}}{\partial \theta} &= (1 - F_{\bm{r}}(t, \theta)) \tfrac{\partial}{\partial \theta} \left( \int_{t_n}^{t} \sigma_{r}(s, \theta) \mathrm d s \right). \label{eq:dfdtheta}
\end{align}
We solve Eq.~\ref{eq:differential} for $\tfrac{\partial t}{\partial \theta}$ and substitute the partial derivatives using Eqs.~\ref{eq:dfdt} and~\ref{eq:dfdtheta} to obtain the final expression for the gradient
\begin{equation}\label{eq:t_gradient}
\frac{\partial t}{\partial \theta} = \frac{\tfrac{\partial y}{\partial \theta} - (1 - F_{\bm{r}}(t, \theta)) \tfrac{\partial}{\partial \theta} \int_{t_n}^t \sigma_{\bm{r}}(s,\theta) \mathrm d s}{ (1 - F_{\bm{r}}(t, \theta)) \sigma_{\bm{r}}(t, \theta)}.
\end{equation}
Automatic differentiation can be used to compute $\partial y / \partial \theta$ and $\tfrac{\partial}{\partial \theta} \int_{t_n}^{t} \sigma(s) \mathrm d s$ to combine the results as in Eq.~\ref{eq:t_gradient}.

\section{RADIANCE ESTIMATES FOR A SINGLE RAY}\label{sec:toy_exp}

In this section, we evaluate the proposed Monte Carlo radiance estimate (see Eq.~\ref{eq:reparameterized_mc_color_approximation}) in a one-dimensional setting. In this experiment, we assume that we know density in advance and show how the estimate variance depends on the number of radiance calls. Compared to sampling approaches, the standard approximation from Eq.~\ref{eq:grid_color_approximation} has zero variance but does not allow controlling the number of radiance calls.

\begin{figure}
\centering
\includegraphics[width=0.7\linewidth]{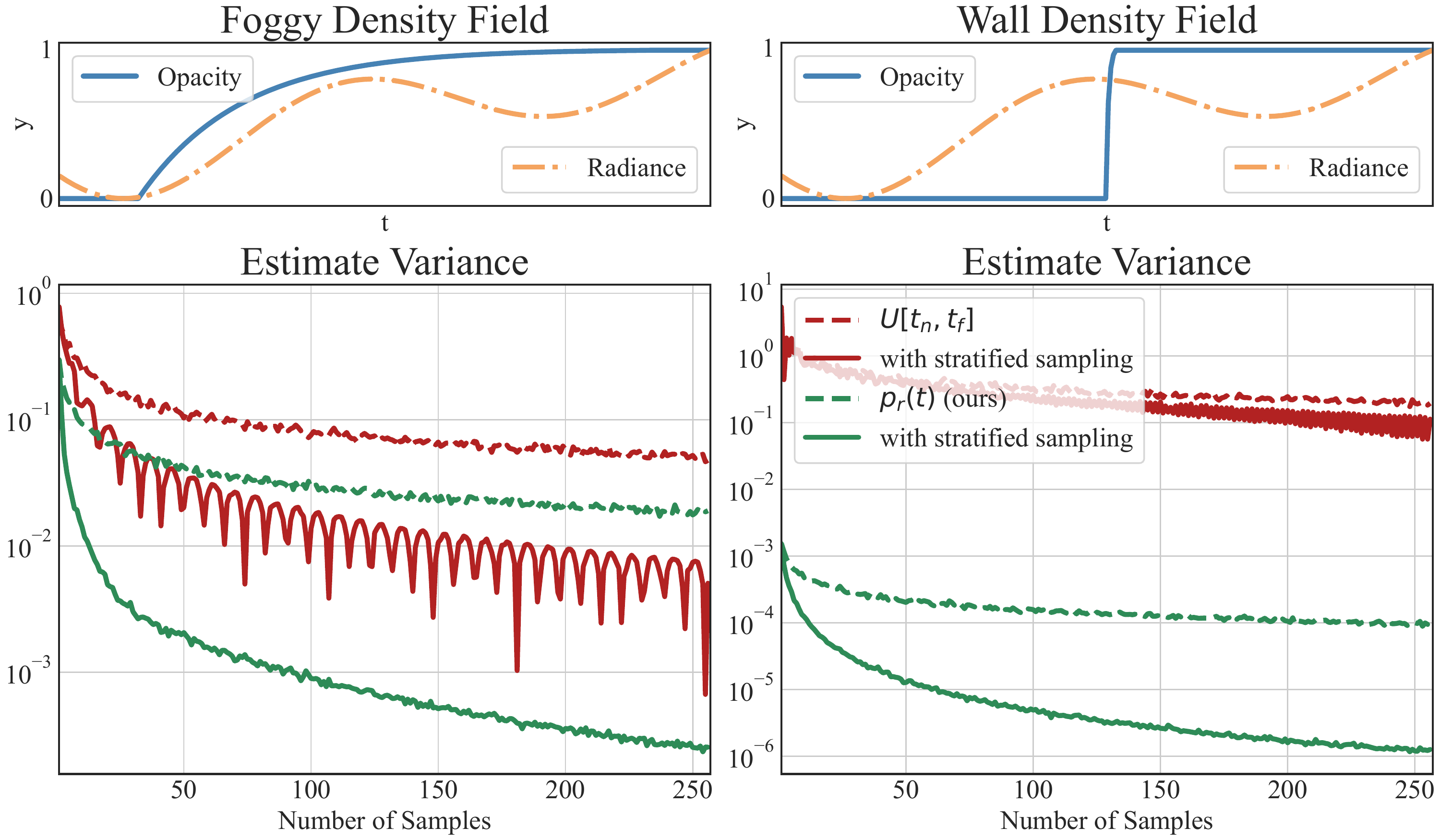}
\caption{Color estimate variance compared for a varying number of samples. The upper plot illustrates underlying opacity function on a ray; the lower graph depicts variance in logarithmic scale.
Compared to a naive estimate of the integral with uniform samples (dashed red), inverse transform sampling exhibits lower variance (dashed green). Stratified sampling improves variance in both setups (solid lines).}
\label{fig:toy_experiment}
\end{figure}

Our experiment models light propagation on a single ray in two typical situations. The upper row of Fig.~\ref{fig:toy_experiment} defines a scalar radiance field (orange) $c_{\bm{r}}(t)$ and opacity functions (blue) $F_{\bm{r}}(t)$ for \textit{"Foggy"} and \textit{"Wall"} density fields. The first models a semi-transparent volume, which often occurs after model initialization during training. In the second, light is emitted from a single point on a ray, which is common in applications.

For the two fields we estimated the expected radiance $C(\bm{r}) = \int_{t_n}^{t_f} c_{\bm{r}}(t) \mathrm d F_{\bm{r}}(t)$. We consider two baseline methods (both in red in Fig.~\ref{fig:toy_experiment}): the first is a Monte Carlo estimate of $C$ obtained with uniform distribution on a ray $U[t_n, t_f]$, and its stratified modification with a uniform grid $t_n = t_0 < \dots < t_k = t_f$ (note that here we use $k$ to denote the number of samples, not the number of grid points $m$ in piecewise density approximation):
\begin{equation}\label{eq:stratified_mc_grid_approximation}
\hat{C}_{\text{IW}}(\bm{r}) = \sum_{i=1}^k (t_i - t_{i-1}) c_{\bm{r}}(\tau_i) \frac{\mathrm d F_{\bm{r}}}{\mathrm d t} \bigg\rvert_{t = \tau_i},
\end{equation}
where $\tau_i \sim U[t_{i - 1}, t_i]$ are independent uniform bin samples.
We compare the baseline against the estimate from Eq.~\ref{eq:reparameterized_mc_color_approximation} and its stratified modification. All estimates are unbiased. Therefore, we only compare the estimates' variances for a varying number of samples $m$.

In all setups, our stratified estimate uniformly outperforms the baselines. For the more challenging "foggy" field, approximately $k=32$ samples are required to match the baseline performance for $k=256$. We match the baseline with only a $k = 4$ samples for the "wall" field. Inverse transform sampling requires only a few points for degenerate distributions. 

\newpage

\section{COMPUTATIONAL EFFICIENCY OF APPROXIMATIONS}\label{appendix:computational_efficiency}
In this section, we analyze our approximation method's computational efficiency and compare it with the numerical approximation from Eq.~\ref{eq:grid_color_approximation}. To illustrate it in a more practical setting, we determine the time performance of the estimates relative to the batch processing time of a radiance field model. As an example, we pick a recent voxel-based radiance field model DVGO~\cite{sun2022direct} that parameterizes density field as a voxel grid and combines voxel grid with a view-dependent neural network to obtain a hybrid parameterization of radiance.

\begin{table}[H]
\centering
\caption{Computational complexity of three stages of color estimation.}
\begin{tabular}{l|ccc}
                  & Voxel Grid ($\sigma$)& Hybrid ($c$) & $\mathbb{E}\hat{C}$\\
\hline
Baseline & 0.00025s & 0.02887s & 0.00187s \\
Ours, 32 & 0.00025s & 0.00419s & 0.00369s \\
Ours, 64 & 0.00025s & 0.00665s & 0.00372s \\
Ours, 128 & 0.00025s & 0.01334s & 0.00381s \\
Ours, 256 & 0.00025s & 0.02887s & 0.00383s 
\end{tabular}
\label{tab:compute} 
\end{table}

Despite the toy nature of the experiment, we focus on a setting and hyperparameters commonly used in the rendering experiments. We take a batch of size 2048, draw 256 points along each of the corresponding rays and use them to calculate density $\sigma$. In the baseline setting, we then use the same points to calculate radiance $c$ and estimate pixel colors with Eq.~\ref{eq:grid_color_approximation}. In contrast to this pipeline, we propose to use calculated values of $\sigma$ to make a piecewise constant approximation of density, generate a varying number (32, 64, 128, 256) of Monte Carlo samples and use them to calculate the stratified version of the color estimate (Eq~\ref{eq:reparameterized_mc_color_approximation}). We parameterize $\sigma$ with a voxel grid and $c$ with a hybrid architecture used in DVGO with the default parameters.

In Table~\ref{tab:compute} we report time measurements for each of the stages of color estimation: calculating density field in 256 points along each ray, calculating radiance field in the Monte Carlo samples (or in the same 256 points in case of baseline) and sampling combined with calculating the approximation given $\sigma$ and $c$ (just calculating the approximation in case of baseline). All calculations were made on NVIDIA GeForce RTX 3090 Ti GPU and include both forward and backward passes.

First of all, $\sigma$ computation time is equal for all of the cases and has order of $10^{-4}$ seconds, negligible in comparison with other stages. In terms of calculating the approximation, both baseline and our method work proportionally to $10^{-3}$ seconds, but Monte Carlo estimate take 2.3 to 2.4 times more. Nevertheless, in the mentioned practical scenario this difference is not crucial, since computation of $c$, the heaviest part, takes up to $3 \times 10^{-2}$ seconds. Even in the case of 256 radiance evaluations the difference in total computation time is less than $10\%$. This makes our method at least comparable with the baseline for architectures that can evaluate the density field faster than the radiance field. At the same time, our approach allows to explicitly control the number of radiance evaluations $k$, improving the computational efficiency even further given suitable architectures.

\section{PICKING FINE POINTS IN DIFFERENTIABLE HIERARCHICAL SAMPLING}

\label{sec:hierarchical_details}

\begin{table}[H]
\caption{Comparison of various hierarchical sampling configurations on the Lego scene of Blender dataset.}
\centering
\begin{tabular}{l|cc|ccc|ccc}
 & \multicolumn{2}{c|}{Evaluations} &  \multicolumn{3}{c|}{NeRF} & \multicolumn{3}{c}{RVS} \\
& $N_p$ & $N_f$  & PSNR $\uparrow$ & SSIM $\uparrow$ & LPIPS & PSNR $\uparrow$ & SSIM $\uparrow$ & LPIPS \\
\hline
Union      & 16     & 32   & 25.81     & 0.890  & 0.150 & 27.18     & 0.903  & 0.145 \\
No union      & 16     & 32  & \textbf{27.09}     & \textbf{0.913}  & \textbf{0.121} & \textbf{29.18}     & \textbf{0.928}  & \textbf{0.112} \\
\hline
Union     & 32     & 64  & 29.61     & 0.939  & 0.084 & 30.34     & 0.942  & 0.088 \\
No union     & 32     & 64  & \textbf{30.11}     & \textbf{0.947}  & \textbf{0.070} & \textbf{31.89}     & \textbf{0.955}  & \textbf{0.066} \\
\hline
Union     & 64     & 128   & \textbf{32.14}    & \textbf{0.958}  & 0.053 & 31.63    & 0.952  & 0.068 \\
No union     & 64     & 128   & 31.87    & 0.958  & 0.054 & \textbf{32.80}    & \textbf{0.963}  & \textbf{0.051} \\
\hline
Union     & 64     & 192    & \textbf{32.69}    & \textbf{0.962}  & \textbf{0.048} & 32.46     & 0.960  & 0.056 \\
No union      & 64     & 192  & 32.14    & 0.960  & 0.051 & \textbf{33.03}     & \textbf{0.964}  & \textbf{0.047} \\
\end{tabular}
\label{tab:nerf_lego_uni} 
\end{table}

We consider two options for picking $N_f$ points for fine network evaluation: either take $N_f$ samples from the proposal distribution (first option), or take the union of $N_p$ grid points that were used to construct the distribution and $N_f - N_p$ new samples from the proposal distribution (second option, originally used in NeRF). In Table~\ref{tab:nerf_lego}, we report the best result out of the two options, and Table~\ref{tab:nerf_lego_uni} presents the results for both options. Our method works best with the first option. The baseline performs better with the first option for $(16, 32)$, $(32, 64)$ configurations and with the second option for other configurations. Comparisons in Table~\ref{tab:nerf_blender_llff} are done in $(N_p = 32, N_f = 64)$ configuration with the first option. We use this configuration for comparison as both methods perform best with the same option and produce better results than in $(N_p = 16, N_f = 32)$ configuration. In the ablation study in Table~\ref{tab:nerf_ablation}, we also show that the base model works worse with the second option in $(N_p = 32, N_f = 64)$ configuration on average across other scenes from Blender dataset.

\section{ADDITIONAL VISUALIZATIONS}
\label{appendix:visualizations}

\begin{figure}[H]
\centering
\includegraphics[width=0.99\linewidth]{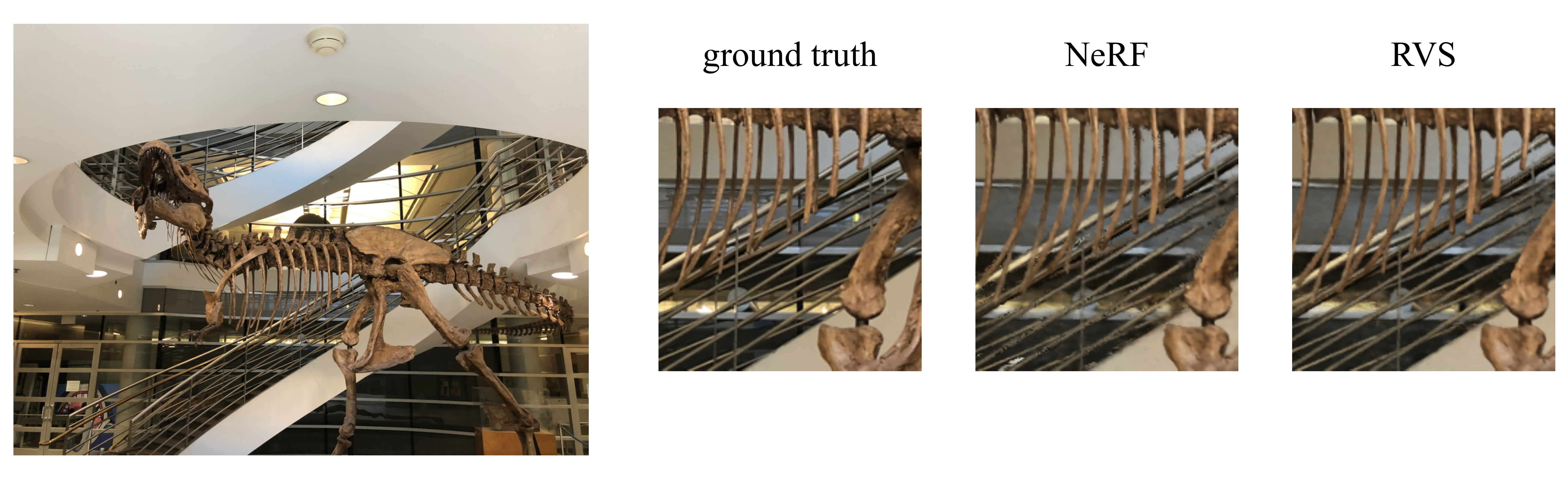}
\caption{Comparison between renderings of a test-set view on T-Rex scene (LLFF) for $(32, 64)$ configuration. Artifacts can be seen on railings and bones in the NeRF render, while such artifacts are absent in the reconstruction produced by our model.}
\label{fig:nerf_trex}
\end{figure}

\begin{figure}[H]
\centering
\includegraphics[width=1.0\linewidth]{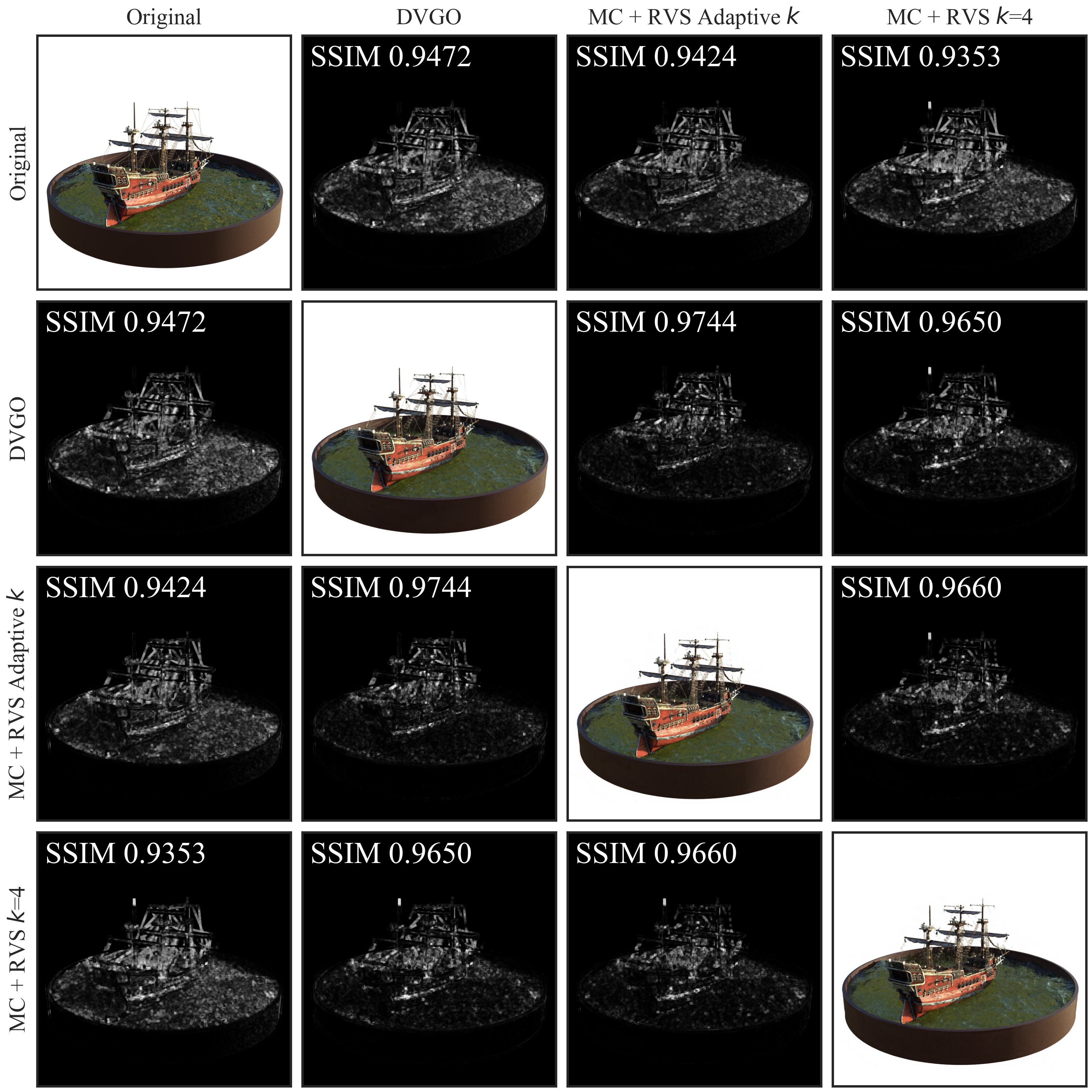}
\caption{Qualitative comparison of scene reconstruction based on various training approaches. We visualize the default DVGO model along with two models trained using the reparameterized volume sampling algorithm (with the adaptive number of samples and $k=4$ samples). In all three models, the novel view does noes faithfully capture the water surface and fine ship details. Noticeably, the three model align in water surface reconstruction as opposed to the ship details. Perceptual quality of the details follows the numerical results.}
\end{figure}

\begin{figure}[H]
\centering
\includegraphics[width=1.0\linewidth]{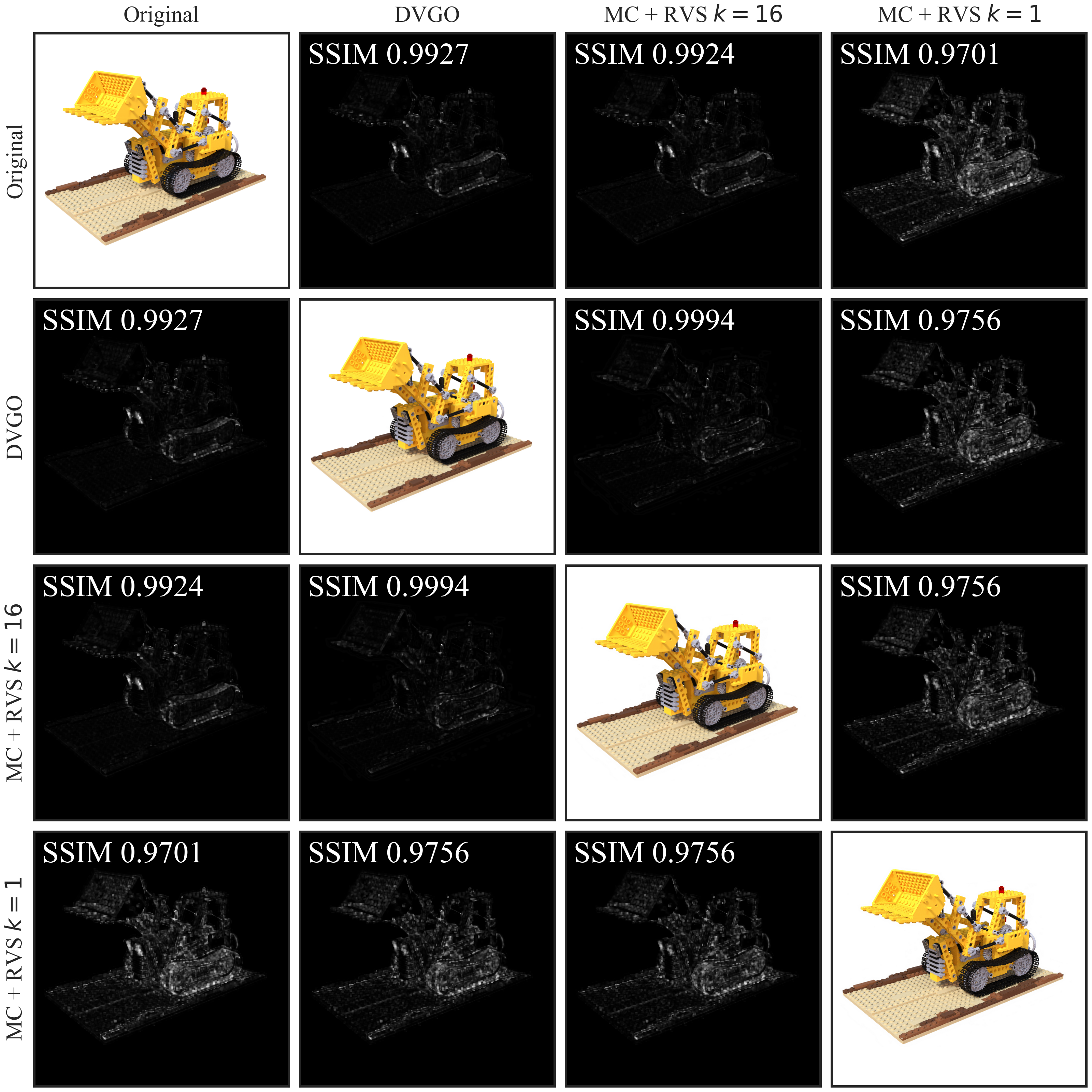}
\caption{Comparison of novel views produced by different rendering algorithms. We train a base DVGO model and render novel views using the default rendering algorithm and the proposed algorithm with $k=1, 16$ samples. Except for a single-sample estimate, the rendered images on diagonal are barely distinguishable. Outside the diagonal, we visualize the difference with SSIM gradients and report SSIM between the images. The images produced with the default rendering algorithm and our algorithm with $k=16$ differ the least, whereas the image with $k$ samples has noticeable flaws near the track and edges.}
\end{figure}

\section{PER-SCENE RESULTS}
\label{appendix:per-scene-results}

\begin{table}[H]
\centering
\small
\caption{Differentiable hierarchical sampling with NeRF on Blender dataset.}
\resizebox{\textwidth}{!}{%
\begin{tabular}{l|cccccccc|c}
PSNR$\uparrow$                             & Chair & Drums & Ficus & Hotdog & Lego  & Materials & Mic   & Ship  & Avg   \\
\hline
NeRF, $(N_p = 32, N_f = 64)$                                       & 31.33 & 23.89 & 28.26 & 35.44  & 30.11 & 28.65 & 31.46 & 26.76 & 29.49 \\
RVS, $(N_p = 32, N_f = 64)$                              & 31.99 & 24.60 & 29.27 & 36.18  & 31.89 & 29.31     & 31.84 & 27.19 & {30.26} \\
\end{tabular}}
\resizebox{\textwidth}{!}{%
\begin{tabular}{l|cccccccc|c}
SSIM$\uparrow$                             & Chair & Drums & Ficus & Hotdog & Lego  & Materials & Mic   & Ship  & Avg   \\
\hline                            
NeRF, $(N_p = 32, N_f = 64)$                                       & 0.956 & 0.908 & 0.949 & 0.972 & 0.947 & 0.940 & 0.973 & 0.834 & 0.934 \\
RVS, $(N_p = 32, N_f = 64)$                              
& 0.958 & 0.915 & 0.955 & 0.974 & 0.955 & 0.946 & 0.974 & 0.834 & {0.939} \\           
\end{tabular}}
\resizebox{\textwidth}{!}{%
\begin{tabular}{l|cccccccc|c}
LPIPS$\downarrow$                             & Chair & Drums & Ficus & Hotdog & Lego  & Materials & Mic   & Ship  & Avg   \\
\hline                            
NeRF, $(N_p = 32, N_f = 64)$                                       & 0.059 & 0.116 & 0.063 & 0.050 & 0.070 & 0.072 & 0.033 & 0.220 & 0.085 \\
RVS, $(N_p = 32, N_f = 64)$                              
& 0.056 & 0.113 & 0.060 & 0.050 & 0.066 & 0.065 & 0.033 & 0.219 & {0.082} \\              
\end{tabular}}
\label{tab:nerf-blender-all} 
\end{table}

\begin{table}[H]
\centering
\small
\caption{Differentiable hierarchical sampling with NeRF on LLFF dataset.}
\resizebox{\textwidth}{!}{%
\begin{tabular}{l|cccccccc|c}
PSNR$\uparrow$  & Room & Fern & Leaves & Fortress & Orchids  & Flower & T-Rex  & Horns  & Avg   \\
\hline
NeRF, $(N_p = 32, N_f = 64)$                                       & 31.22 & 24.79 & 20.81 & 31.00 & 20.32 & 27.41 & 25.89 & 26.64 & 26.01 \\
RVS, $(N_p = 32, N_f = 64)$                              & 31.87 & 24.89 & 20.89 & 31.11 & 20.27 & 27.40 & 26.48 & 26.98 & {26.24} \\
\end{tabular}}
\resizebox{\textwidth}{!}{%
\begin{tabular}{l|cccccccc|c}
SSIM$\uparrow$ & Room & Fern & Leaves & Fortress & Orchids  & Flower & T-Rex  & Horns  & Avg   \\
\hline                            
NeRF, $(N_p = 32, N_f = 64)$                                       & 0.938 & 0.771 & 0.678 & 0.875 & 0.630 & 0.822 & 0.858 & 0.798 & 0.796 \\
RVS, $(N_p = 32, N_f = 64)$                              
& 0.942 & 0.773 & 0.681 & 0.878 & 0.630 & 0.823 & 0.869 & 0.795 & {0.799} \\           
\end{tabular}}
\resizebox{\textwidth}{!}{%
\begin{tabular}{l|cccccccc|c}
LPIPS$\downarrow$  & Room & Fern & Leaves & Fortress & Orchids  & Flower & T-Rex  & Horns  & Avg   \\
\hline                            
NeRF, $(N_p = 32, N_f = 64)$                                       & 0.203 & 0.309 & 0.328 & 0.187 & 0.338 & 0.226 & 0.279 & 0.310 & 0.273 \\
RVS, $(N_p = 32, N_f = 64)$                              
& 0.193 & 0.310 & 0.328 & 0.182 & 0.340 & 0.223 & 0.267 & 0.311 & {0.270} \\              
\end{tabular}}
\label{tab:nerf-llff-all} 
\end{table}

\begin{table}[H]
\centering
\caption{Differentiable hierarchical sampling with NeRF++ on LF dataset.}
\resizebox{0.7\textwidth}{!}{%
\begin{tabular}{l|cccc|c}
PSNR$\uparrow$  & Africa & Basket & Torch & Ship  & Avg   \\
\hline
NeRF++, $(N_p = 32, N_f = 64)$                                       & 26.36 & 21.38 & 23.72 & 24.53 & 23.99 \\
RVS, $(N_p = 32, N_f = 64)$                              & 27.31 & 21.58 & 24.60 & 25.01 & {24.63} \\
\end{tabular}}
\resizebox{0.7\textwidth}{!}{%
\begin{tabular}{l|cccc|c}
SSIM$\uparrow$ & Africa & Basket & Torch & Ship  & Avg    \\
\hline                            
NeRF++, $(N_p = 32, N_f = 64)$                                       & 0.838 & 0.790 & 0.767 & 0.744 & 0.784 \\
RVS, $(N_p = 32, N_f = 64)$                              
& 0.865 & 0.812 & 0.797 & 0.777 & {0.812} \\           
\end{tabular}}
\resizebox{0.7\textwidth}{!}{%
\begin{tabular}{l|cccc|c}
LPIPS$\downarrow$  & Africa & Basket & Torch & Ship  & Avg    \\
\hline                            
NeRF++, $(N_p = 32, N_f = 64)$                                       & 0.221 & 0.302 & 0.297 & 0.329 & 0.287 \\
RVS, $(N_p = 32, N_f = 64)$                              
& 0.177 & 0.290 & 0.258 & 0.288 & {0.253} \\              
\end{tabular}}
\label{tab:nerfpp-lf-all} 
\end{table}

\begin{table}[H]
\centering
\caption{Differentiable hierarchical sampling with NeRF++ on T\&T dataset.}
\resizebox{0.72\textwidth}{!}{%
\begin{tabular}{l|cccc|c}
PSNR$\uparrow$  & Truck & Train & M60 & Playground  & Avg   \\
\hline
NeRF++, $(N_p = 32, N_f = 64)$                                       & 21.18 & 17.16 & 16.96 & 21.55 & 19.21 \\
RVS, $(N_p = 32, N_f = 64)$                              & 21.62 & 17.32 & 17.48 & 22.05 & {19.62} \\
\end{tabular}}
\resizebox{0.72\textwidth}{!}{%
\begin{tabular}{l|cccc|c}
SSIM$\uparrow$ & Truck & Train & M60 & Playground  & Avg    \\
\hline                            
NeRF++, $(N_p = 32, N_f = 64)$                                       & 0.661 & 0.539 & 0.617 & 0.633 & 0.612 \\
RVS, $(N_p = 32, N_f = 64)$                              
& 0.666 & 0.545 & 0.619 & 0.657 & {0.622} \\           
\end{tabular}}
\resizebox{0.72\textwidth}{!}{%
\begin{tabular}{l|cccc|c}
LPIPS$\downarrow$  & Truck & Train & M60 & Playground  & Avg    \\
\hline                            
NeRF++, $(N_p = 32, N_f = 64)$                                       & 0.423 & 0.541 & 0.516 & 0.493 & 0.493 \\
RVS, $(N_p = 32, N_f = 64)$                              
& 0.410 & 0.527 & 0.506 & 0.446 & {0.472} \\              
\end{tabular}}
\label{tab:nerfpp-tt-all} 
\end{table}

\begin{table}[H]
\caption{DVGO with Monte Carlo estimates on Blender dataset.}
\small
\centering
\resizebox{0.95\textwidth}{!}{%
\begin{tabular}{l|cccccccc|c}
PSNR$\uparrow$                             & Chair & Drums & Ficus & Hotdog & Lego  & Materials & Mic   & Ship  & Avg   \\
\hline
DVGO                                       & 34.07 & 25.40 & 32.59 & 36.75  & 34.64 & 29.58     & 33.14 & 29.02 & 31.90 \\
MC + RVS, $k=4$                                 & 33.79 & 25.16 & 31.81 & 36.22  & 33.35 & 28.32     & 33.03 & 27.87 & 31.19 \\
MC + RVS, $k=8$                                 & 33.49 & 25.16 & 31.30 & 36.26  & 33.34 & 28.64     & 32.87 & 27.99 & 31.13 \\
MC + RVS, \small{adaptive} $k$                  & 34.13 & 25.18 & 31.17 & 36.63  & 33.85 & 29.09     & 33.03 & 28.43 & 31.44
\end{tabular}}
\resizebox{0.95\textwidth}{!}{%
\begin{tabular}{l|cccccccc|c}
SSIM$\uparrow$                             & Chair & Drums & Ficus & Hotdog & Lego  & Materials & Mic   & Ship  & Avg   \\
\hline  
DVGO                                       & 0.976 & 0.929 & 0.977 & 0.980 & 0.976 & 0.950 & 0.983 & 0.877 & 0.956 \\
MC + RVS, $k=4$                                 & 0.976 & 0.927 & 0.975 & 0.979 & 0.971 & 0.938 & 0.982 & 0.857 & 0.951 \\
MC + RVS, $k=8$                                 & 0.973 & 0.927 & 0.972 & 0.979 & 0.970 & 0.942 & 0.981 & 0.861 & 0.951 \\
MC + RVS, \small{adaptive} $k$                  & 0.977 & 0.928 & 0.972 & 0.980 & 0.973 & 0.946 & 0.982 & 0.870 & 0.953               
\end{tabular}}
\resizebox{0.95\textwidth}{!}{%
\begin{tabular}{l|cccccccc|c}
LPIPS$\downarrow$                          & Chair & Drums & Ficus & Hotdog & Lego  & Materials & Mic   & Ship  & Avg   \\
\hline  
DVGO                                       & 0.028 & 0.080 & 0.025 & 0.034  & 0.027 & 0.058     & 0.018 & 0.162 & 0.054 \\
MC + RVS, $k=4$                                 & 0.028 & 0.079 & 0.028 & 0.038  & 0.032 & 0.070     & 0.017 & 0.181 & 0.059 \\
MC + RVS, $k=8$                                 & 0.031 & 0.080 & 0.030 & 0.037  & 0.033 & 0.067     & 0.019 & 0.178 & 0.059 \\
MC + RVS, \small{adaptive} $k$                  & 0.027 & 0.081 & 0.031 & 0.034  & 0.031 & 0.062     & 0.019 & 0.165 & 0.056
\end{tabular}}
\end{table}

\end{document}